\documentclass[preprint,12pt,authoryear]{elsarticle}

\usepackage{amssymb}

\usepackage{newtxmath, algorithm}
\usepackage[noend]{algpseudocode}
\usepackage{caption, subcaption}
\usepackage{tabularx, makecell, booktabs, siunitx}
\setcitestyle{square,sort,comma,numbers}

\journal{Robotics and Autonomous Systems}

\begin{document}
\captionsetup[figure]{labelsep=period,name={Fig.}, labelfont=bf}
\captionsetup[table]{labelsep=newline, labelfont=bf}
\setcitestyle{open={[},close={]}}

\begin{frontmatter}



\title{A Versatile Door Opening System with Mobile Manipulator through Adaptive Position-Force Control and Reinforcement Learning }

\author[a]{Gyuree Kang}
\ead{fingb20@kaist.ac.kr}
\author[a]{Hyunki Seong}
\ead{hynkis@kaist.ac.kr}
\author[a]{Daegyu Lee}
\ead{lee.dk@kaist.ac.kr}
\author[a]{D.Hyunchul Shim\corref{cor1}}
\ead{hcshim@kaist.ac.kr}
\cortext[cor1]{Corresponding author}
\affiliation[a]{organization={School of Electrical Engineering},
            state={Daejeon},
            country={Republic of Korea}}

\begin{abstract}
The ability of robots to navigate through doors is crucial for their effective operation in indoor environments.
Consequently, extensive research has been conducted to develop robots capable of opening specific doors.
However, the diverse combinations of door handles and opening directions necessitate a more versatile door opening system for robots to successfully operate in real-world environments.
In this paper, we propose a mobile manipulator system that can autonomously open various doors without prior knowledge. 
By using convolutional neural networks, point cloud extraction techniques, and external force measurements during exploratory motion, we obtained information regarding handle types, poses, and door characteristics.
Through two different approaches, adaptive position-force control and deep reinforcement learning, we successfully opened doors without precise trajectory or excessive external force.
The adaptive position-force control method involves moving the end-effector in the direction of the door opening while responding compliantly to external forces, ensuring safety and manipulator workspace.
Meanwhile, the deep reinforcement learning policy minimizes applied forces and eliminates unnecessary movements, enabling stable operation across doors with different poses and widths.
The RL-based approach outperforms the adaptive position-force control method in terms of compensating for external forces, ensuring smooth motion, and achieving efficient speed. It reduces the maximum force required by 3.27 times and improves motion smoothness by 1.82 times. 
However, the non-learning-based adaptive position-force control method demonstrates more versatility in opening a wider range of doors, encompassing revolute doors with four distinct opening directions and varying widths.
\end{abstract}




\end{frontmatter}

\section{Introduction}
\label{sec:introduction}
\begin{figure}[h!]
\centering
\includegraphics[width=0.8\columnwidth]{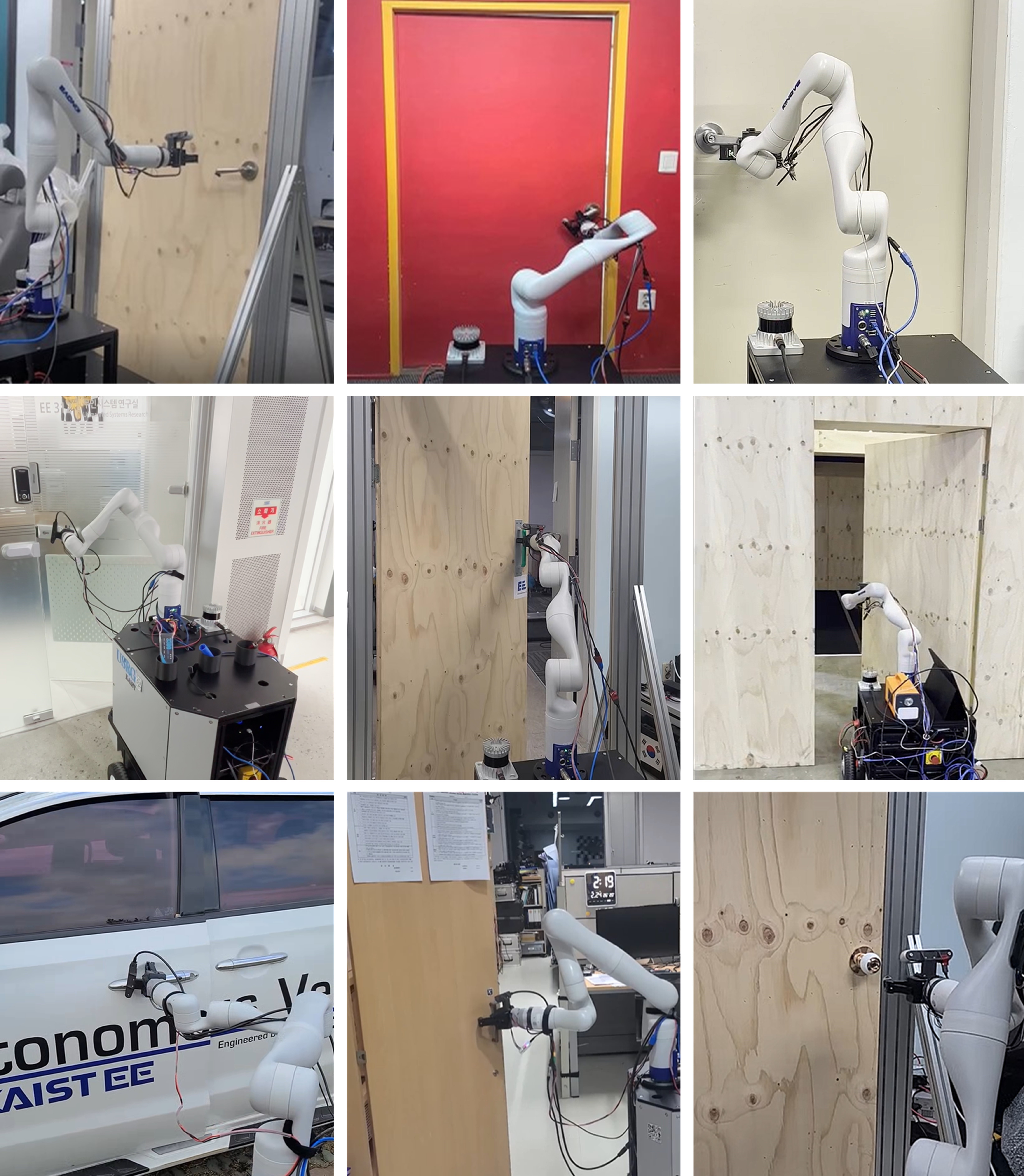}
\caption{Mobile manipulator opening various doors.} \label{fig:figures/various_doors_robot}
\end{figure}
With the increasing use of robots in various aspects of daily life, their working environments are expanding beyond well-defined spaces to include unstructured settings such as office buildings and streets.
Therefore, their ability to interact effectively with their surroundings has become essential.
While autonomous mobile robots have found applications in human-centered fields such as delivery services \cite{lee2021assistive}, \cite{srinivas2022autonomous}, their accessibility to human-oriented facilities, particularly doors, remains a challenge.
Hence, there have been numerous studies on door opening using mobile manipulators. 
However, many of these studies have focused on investigating approaches for opening a certain type of door or examining isolated part of the door opening process, such as handle detection and pose estimation, handle unlatching, or door opening.
Additionally, the validation in real-world environments is often lacking, as most of the evaluations have been conducted solely in simulations.
Therefore, in this study, we propose a system utilizing a mobile manipulator to perform the complete process of door opening and passage.
Our approach aims to address the limitations by considering various types of door handles, opening directions, and door widths and evaluate them in the real-world.

The challenge of opening doors arises from the kinematically constrained situation of the robot while holding the door.
The lack of compliance in the robots' hardware can lead to dangerous levels of force being exerted on both the door and the robot even with slight deviations in their movements.
To address this challenge, the robot must understand the kinematic constraints of the target door, including its rotation axis, opening direction, and the kinematics of the handles.
\cite{peterson2000high} utilized online estimation of the door radius to enable a mobile manipulator to perform rotational door opening movements.
They employed a hybrid dynamic system that incorporated compliant motion, door model predictive control, and ideal manipulator configuration control.
\cite{ahmad2012multiple}  adopted multiple working modes for the joints to prevent excessive internal forces during the target door opening.
They switched the joints parallel to the hinge to passive mode and employed a door parameter estimation method based on recorded motion to generate the end-effector trajectory.
While these studies estimated the target door's radius online to generate the robot trajectory, allowing the system to be applicable to different door types, they only tested the system on pulling-type doors and did not provide a solution to other parts of the door opening process.
\cite{ma2018optimal} focused on optimizing the energy consumption during the door opening motion.
However, their manipulator trajectory was generated based on the measured door radius, making the system dependent on the accuracy of sensor measurements.
Meanwhile, \cite{stuede2019door} adopted impedance control and convolutional neural networks (CNN) for handle detection in opening a door with a mobile manipulator.
They also considered the robot's traversability while opening the door and were able to unlatch handles with different shapes.
However, their system was only deployed on a single type of handle and a pushing door.
\cite{arduengo2021robust} also used CNN to detect the target handle and door, while estimating the pose with a random sample consensus (RANSAC) algorithm.
Then, they estimated door kinematics for prismatic and revolute type doors based on the Bayesian approach and utilized human demonstration as prior knowledge.
This approach allowed the system to unlatch different types of handles but did not consider passing the door.

However, designing planning and control systems for highly complex systems such as mobile manipulators is a challenge for conventional methods, as they are sensitive to parameters and require significant engineering effort \cite{minniti2019whole}.
To address this issue, reinforcement learning (RL) has emerged as a promising approach for complex robot control  \cite{luo2019reinforcement}-\cite{beltran2020variable} including mobile manipulators \cite{wang2020learning}-\cite{sun2022fully}.
As it can be less sensitive to hardware calibration \cite{liu2021deep}, generate robot actions directly from low-level sensor observations, and learn the policy by itself without much prior knowledge and fine-tuning \cite{ibarz2021train}, it is suitable for tasks such as door opening.
\cite{gu2020deep} developed a learning-based door opening robot, where they adopted asynchronous NAF(Normalized Advantage Function) to train the network with experience collected by multiple robots in the real-world.
This enabled faster learning of the robot for complex manipulation in the real-world application without prior demonstrations.
Meanwhile, \cite{nemec2017door} sped up the training of a manipulator for door opening by using physical constraints of the task as a guideline for robot motion planning.
The intelligent control algorithm used in searching RL policy parameters and the training result was tested both in simulation and in the real-world.
Both studies successfully implemented the agent in the real-world but used a fixed manipulator to open doors.
\cite{Door_gym} took the domain randomization problem for the door opening task a step further by introducing a simulation environment with different door parameters and robot properties. 
They also tested the simulation by training the agent with PPO and SAC algorithms, which showed a success rate of up to 0.95 when tested in a randomized environment.
However, they focused only on the door opening without integration into the door-passing task.
\cite{wang2022research} showed door opening with a mobile manipulator, where RL was used for generating the door opening motion.
They used an improved PPO algorithm for training but only focused on a single type of door and handle without utilizing the mobile part of the robot to pass through or approach the door. 
Other than RL, \cite{scirobotics} used deep predictive learning (DPL) to open and pass through the door with the mobile manipulator. 
The task was divided into three parts: approaching, opening, and passing, while each part was executed with the proposed motion generation module. 
The modules generated predictive images and robot motions for the near future, and the one with the lowest standardization error was chosen for the final execution.
However, their study did not demonstrate generalization ability in terms of door type, showing single door opening case only.

In this paper, we propose a versatile door opening system capable of opening doors with various handle types and opening directions.
The system consists of a perception module that estimates the class and pose of the target handle from an RGBD image using a segmentation algorithm.
Additionally, the robot estimates door kinematics through exploratory motion and force measurement with a force/torque sensor at the end-effector.
During the door opening and passing phase, we implement and compare non-learning adaptive position-force control and RL-based techniques, as they are two of the most extensively employed methods for adaptive door opening.
The adaptive position-force control method, known for its ease of adjustability and adaptability to different door types, proved to be less sensitive to environmental variations.
Leveraging information from preceding modules, it successfully opened doors with different handle types, opening directions, and widths.
In contrast, the RL method demonstrated its effectiveness in achieving diverse objectives, considering intricate robot actions, and utilizing various sensor inputs by exhibiting superior performance in terms of compliance, stability, and speed when dealing with pushing doors of varying widths.

The contributions of this study are as follows.
\begin{itemize}
\item Proposing versatile full sequence door opening system framework for mobile manipulators, applicable to doors with various handles and kinematic constraints without prior knowledge.
\item Designing both adaptive position-force-based system and RL-based hybrid system for adaptive door opening.
\item Implementing both systems and extensively evaluating them in real-world scenarios.
\end{itemize}
 
The rest of the paper is organized as follows:
In Section \ref{sec:task_planning}, we provide a brief overview of the four subtask modules, followed by a detailed description of each module in Sections \ref{sec:detection} - \ref{sec:learning_based}.
In particular, Sections \ref{sec:non_learning_based} and \ref{sec:learning_based} explain the two approaches for adaptive door opening and passing motion.
We describe the experiment settings in Section \ref{sec:experiments}, and present the simulation and real-world test results in Section \ref{sec:results}.
Finally, we provide our conclusions in Section \ref{sec:conclusion}.

\section{Methods}
\label{sec:task_planning}
\begin{figure*}[h!]
\centering
\includegraphics[width=0.8\textwidth]{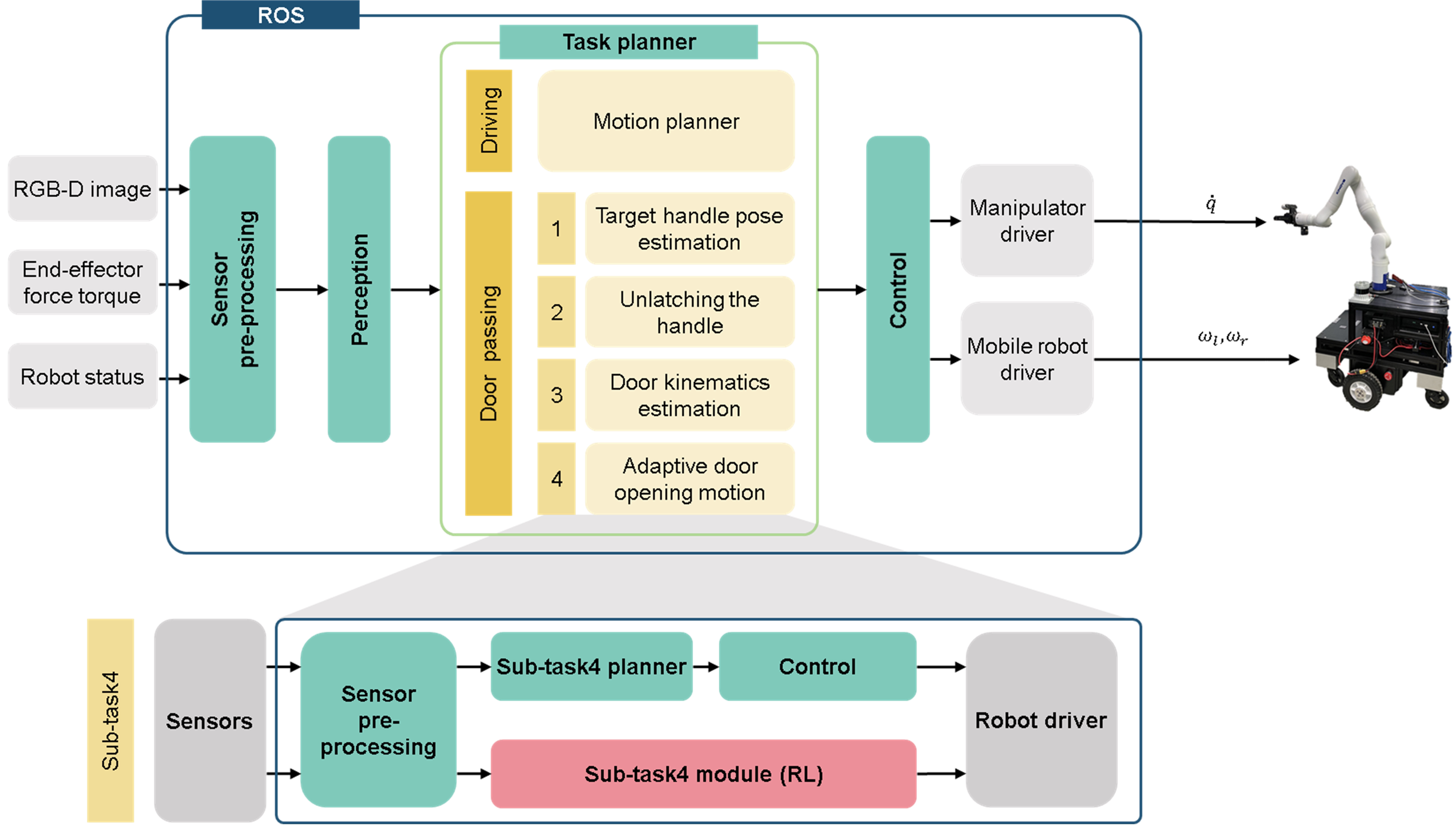}
\caption{Mobile manipulator door opening system architecture. The controller in subtask 4 (ST4) for adaptive door opening motion was made with two different approaches: adaptive position-force controller, and deep reinforcement algorithm.} \label{fig:mm_system}
\end{figure*}

To pass through a door using a mobile manipulator, we divide the task into four subtasks, which are solved with corresponding modules as shown in Fig. \ref{fig:mm_system}.
The first subtask is estimating the pose and the type of the target handle (ST1) and approaching it until it is within reach of the manipulator.
The second subtask is unlatching the handle (ST2) based on the type defined from ST1. 
The third subtask is identifying the opening direction of the door (ST3).
The fourth subtask is opening and passing through the door (ST4) according to the opening direction identified in ST3 and passing through it without collision.
After the robot finishes navigating through the door, it switches back to autonomous driving mode.

\subsection{Handle pose estimation}
\label{sec:detection}
\begin{figure*}[h!]
\centering
\includegraphics[width=0.8\textwidth]{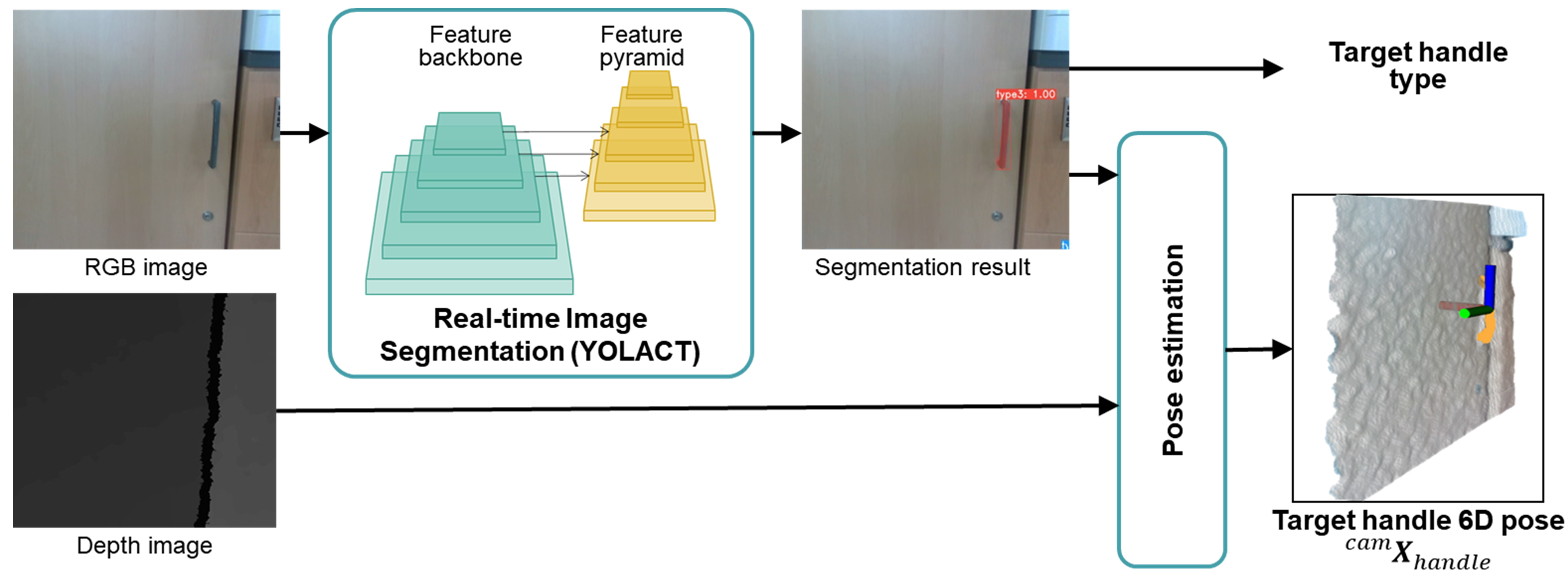}
\caption{Target handle type and pose estimation algorithm. The image segmentation result serves as a mask to filter out the depth image which is subsequently projected to the point cloud of the target handle. The point cloud is utilized to estimate the pose of the target handle.} \label{fig:detection}
\end{figure*}
Since door handles are small objects usually ranging from 20cm to 10cm in width, we combine object detection results from images with depth images to generate point clouds of the target rather than detecting the target directly from the point cloud \cite{lai2022stratified} as in Fig. \ref{fig:detection}.
A real-time image segmentation algorithm \cite{liu2021yolactedge} is utilized to detect and classify the handles into three types: lever type (type 1), door knob (type 2), and bar handle (type 3). 
To generate the point cloud of the target handle from the depth image, the segmented image of the handle is used as a mask $M_h$ to filter out the background from the depth image.
The filtered depth image is then converted into a point cloud ($^{cam}\boldsymbol{P}_h = {p^d_1, p^d_2, ..., p^d_j}$) of the target handle through depth image projection. 

The pose of the handle is estimated using $^{cam}\boldsymbol{P}_h$.
The center of $^{cam}\boldsymbol{P}_h$ is assumed to be the position of the target, whereas the vertical axis from the RANSAC plane segmentation algorithm \cite{fischler1981random} is used to estimate the orientation.
The estimated pose $^{cam}\boldsymbol{X}_{trg}$ of the handle is then transformed to the base frame $^{mb}\boldsymbol{X}_{trg}$ using $T^{mb}_{cam}$, the transformation matrix from the camera frame to the base frame.
Once the pose of the handle has been estimated, the robot approaches $^{mb}\boldsymbol{X}_{trg}$ until it is within reach ($\rho$) of the manipulator, while the manipulator aligns its pose relative to $^{mb}\boldsymbol{X}_{trg}$ in preparation for the handle unlatching motion.
This method of pose estimation is compatible with handles with various shapes, as it does not require CAD or any other prior knowledge of the target object.

\subsection{Unlatch handles}
\label{sec:handle}
\begin{figure}[ht]%
    \centerline{
    \subfloat[]{\includegraphics[height=0.2\columnwidth]{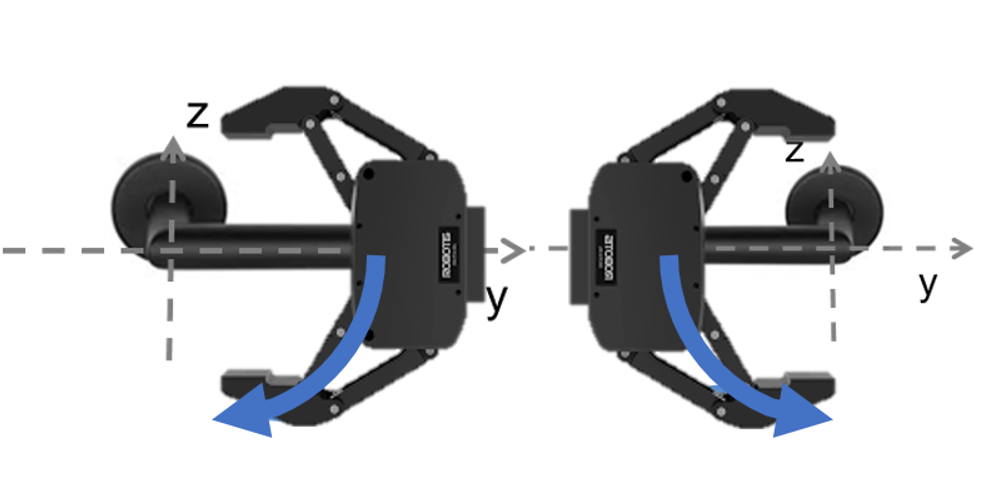}}%
    \subfloat[]{\includegraphics[height=0.2\columnwidth]{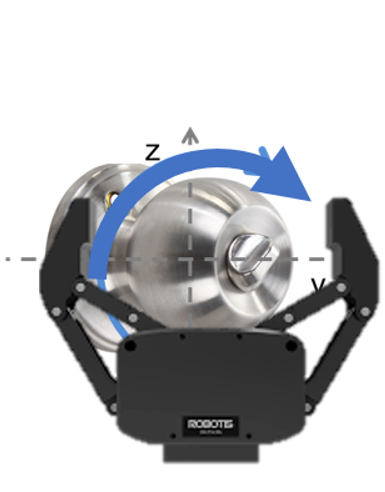}}%
    \subfloat[]{\includegraphics[height=0.2\columnwidth]{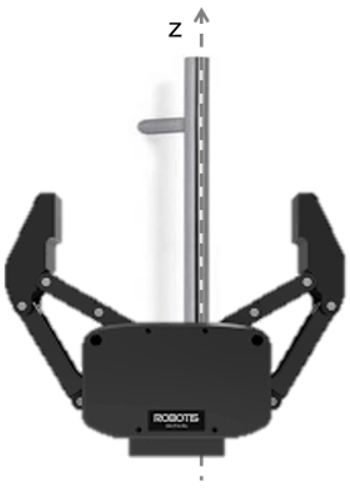}}
}
\caption{Predefined grasping poses according to handle types. (a) type1: levers, (b) type2: doorknobs, and (c) type3: pulling handles.}%
\label{handle_grasp}%
\end{figure}

To unlatch the handle, the robot positions the end-effector to a grasping pose according to the target handle's type as shown in Fig. \ref{handle_grasp}.
To avoid the target getting out of the camera's view and being unable to identify the object when it gets too close, the end-effector slowly reaches out until the handle is detected by the wrist force/torque sensor.
Once the manipulator contacts with the handle, the gripper grasps the target.

\begin{figure}[h]%
\centerline{
    \includegraphics[width=0.3\columnwidth]{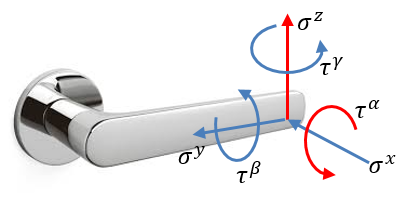}}
\caption[Unlatching motion]{Unlatching handle motion with adaptive position-force control. The red axes represent the manipulation axes, while the blue axes represent the passive axes.}%
\label{unlatching_handle}%
\end{figure}

After grasping the handle, the robot unlatches it using end-effector adaptive position-force control described in Section \ref{sec:non_learning_based}.
The robot exerts force and generates a target position in the manipulation directions of the handle as in Fig. \ref{unlatching_handle}.
The force limits of the manipulation axes for the adaptive position-force control are set high to allow the robot to apply force to unlatch the handle.
Conversely, the limits of the passive (non-manipulation) axes are set low to generate compensating motion for the external force to prevent excessive loading.
The robot continues the unlatching motion until the force/torque of the end-effector reaches its thresholds ($\Omega, \Lambda$) on the manipulation axis.

\subsection{Door kinematics estimation}
\label{sec:kinematics}
The door kinematics comprise the normal opening direction ($\psi^x$) and the horizontal direction ($\psi^{y}$).
Initially, the robot generates the exploratory motion in x-axis to estimate $\psi^x$ by pushing the door.
If the force measurement in the x-axis does not exceed $\Omega^{x}$, the door is considered a push-open door ($\psi^x = 1$).
Otherwise, the robot pulls the door, and if the same condition ($\left|\sigma^x\right| < \Omega^{x}$) is met, the door is defined as a pull-open door ($\psi^x = -1$).

After defining $\psi^x$, the robot opens the door while generating compliant movement to external adaptive position-force control in the passive axis of the door.
If the end-effector moves farther than the threshold in the x-axis ($\delta^{x}$) and relatively left to the previous pose, the horizontal opening direction of the door $\psi^{y}$ is defined as 1; otherwise, it is defined as -1.
The door is considered a clockwise(CW) door if $\psi^x\psi^y$ is 1; otherwise, it is considered a counter-clockwise(CCW) door.

\begin{figure}[h!t]%
\centerline{
    \subfloat[]{\includegraphics[width=0.3\columnwidth]{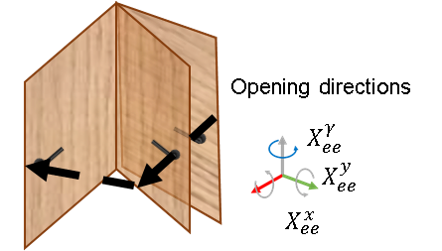}}%
    \subfloat[]{\includegraphics[width=0.3\columnwidth]{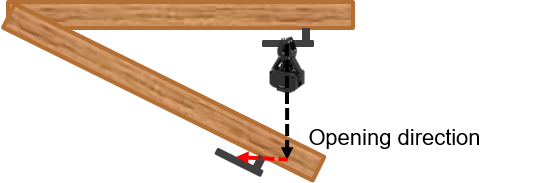}}%
}
\caption{Estimating the opening direction of the door based on exploratory motion and force measurements. (a) motion axis ($x$, $y$, and $\gamma$) for exploratory motion, (b) confirmed opening direction.}%
\label{hinge_location}%
\end{figure}

\subsection{Opening and navigate through doors: adaptive position-force-based method}
\label{sec:non_learning_based}
\begin{figure*}[t]
\centering
\includegraphics[width=0.9\textwidth]{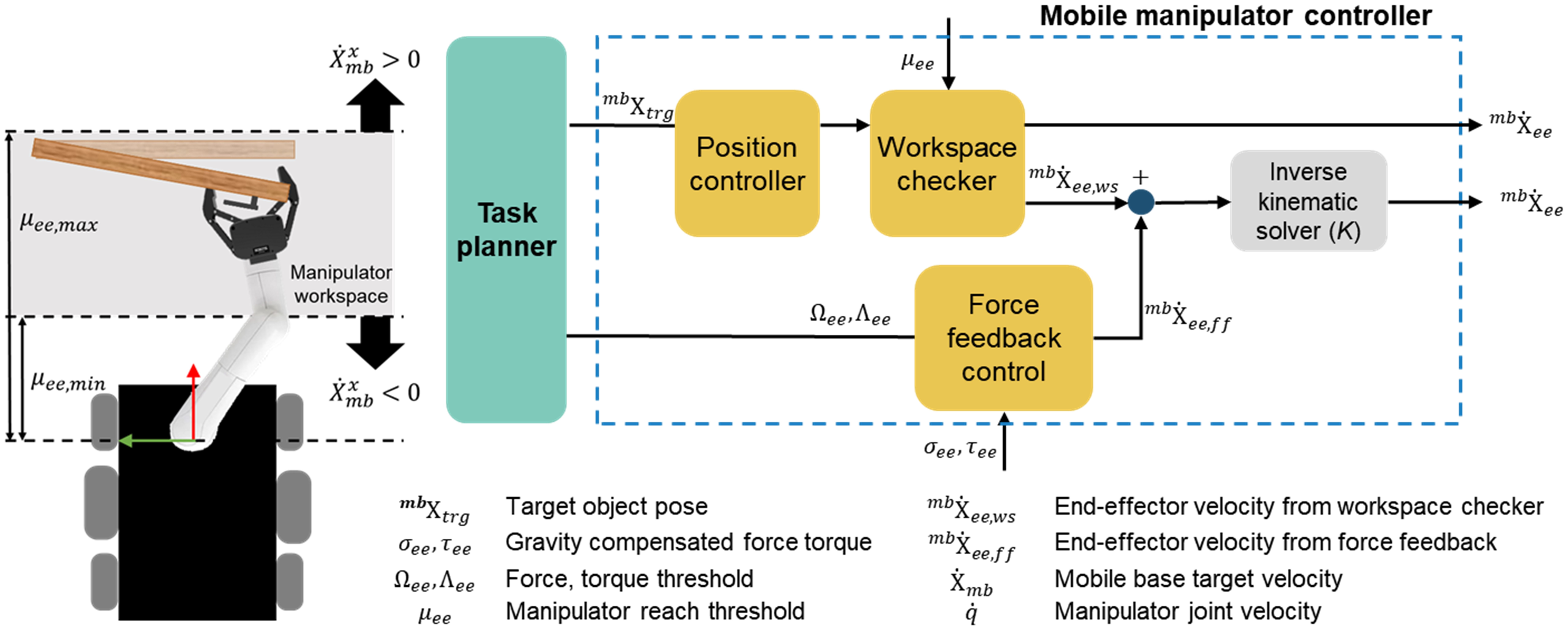}
\caption{The adaptive position-force control system of the mobile manipulator consists of two controllers. The position controller adjusts the end-effector's pose according to the pose of the target object, which is verified by the workspace checker to ensure that it is within the manipulator's reach and the joint position limits. The adaptive position-force controller utilizes force/torque data from the sensor to adaptively control the end-effector's motion in response to external forces.}
\label{fig:mm_control}
\end{figure*}

As depicted in Fig. \ref{fig:mm_control}, the adaptive position-force controller comprises four modules: a position controller, a workspace checker, a force/torque feedback controller, and an inverse kinematic solver (IKS).
Initially, the robot generates the end-effector goal pose towards the opening direction ($\psi^{x}, \psi^{y}$) of the door.
The position controller receives the end-effector's target position $^{mb}{X}_{trg}$ and computes the desired end-effector velocity $^{mb}\dot{X}_{ee,pos}$.

Then workspace checker detects potential collisions between the door and checks if the goal point is within the manipulator's workspace.
If the target point is farther than the workspace of the manipulator ($[\mu_{ee,min}, \mu_{ee,max}]$), the mobile robot moves towards the goal point.
However, if the manipulator and the mobile robot are close enough to predict a potential collision ($\mu_{ee,min}$), the mobile robot moves backward to secure space between them while keeping the end-effector velocity constant.
\begin{equation}
    \label{equ:workspace_checker}
    ^{mb}\dot{X}_{ee,ws} = ^{mb}\dot{X}_{ee,pos} - \dot{X}_{mb}
\end{equation}

The force controller adjusts the target of the end-effector when the gravity-compensated force or torque at the end-effector ($\sigma_{ee}, \tau_{ee}$) exceeds its limits ($\Omega_{ee}$, $\Lambda_{ee}$).
Here, the limits in manipulation axes ($x, y, \gamma$) are set high to enable the robot to apply force to the door while the limits for the passive axes ($z, \alpha, \beta$) are set low to adjust end-effector pose to external forces.
The force controller generates a counter velocity $^{mb}{\dot{X}}_{ee,ff}$ to compensate for the external force as in equation (\ref{PID_compliant_control}).
\begin{equation}
\label{PID_compliant_control}
    ^{mb}{\dot{X}}_{ee,ff} = K_{p,cmp} \sigma_{err}+ K_{i,cmp} \int{\sigma_{err}}dt
\end{equation}
Here, $\sigma_{err}$ is the error between $\sigma_{ee}$ and the force/torque limit ($\Omega_{ee}$, $\Lambda_{ee}$).
Next, the compliant velocity is added to the target velocity using equation (\ref{final_ee_vel}), resulting in the final end-effector velocity $^{mb}\dot{{X}}_{ee}$ and the mobile velocity $\dot{X}_{mb}$.
\begin{equation}
    \label{final_ee_vel}
    ^{mb}\dot{{X}}_{ee} = ^{mb}\dot{{X}}_{ee,ws} + ^{mb}\dot{{X}}_{ee,ff}
\end{equation}
Finally, the joint velocity ($\dot{q} = \mathcal{K}(^{mb}\dot{X}_{ee})$) and wheels' angular velocity ($\omega_l, \omega_r$) are generated from the robot driver when $^{mb}\dot{X}_{ee}$ and $\dot{X}_{mb}$ are given as input.
Here, IKS function $\mathcal{K}$ finds the control values for the corresponding joint velocity $\dot{q}$ given a desired end-effector velocity $^{mb}\dot{X}_{ee}$.

After fully opening the door, the mobile part of the robot moves forward while the manipulator holds the handle to prevent the door from closing.
The gripper then releases the handle and moves to the other side of the door where the base is located. Finally, as the robot moves forward to pass through the door, the manipulator pose is initialized ($q_{init}$), and the task mode changes from door-passing to autonomous driving mode.

\subsection{Opening and navigate through doors: Deep reinforcement learning-based method}
\label{sec:learning_based}

\subsubsection{States and actions}
The robot's state $s_t$ includes the environment and the robot's status.
For the environment state vector, we use distance from the LiDAR point cloud which is the representation of the surrounding obstacles.
To reduce the computational complexity of the input, we use only the closest points located within the front half of the robot in 9-degree intervals($\boldsymbol{S} = \{{s}_{1}, {s}_{2}, ..., {s}_{20}\}$).

For the manipulator's state, we use the measured joints' position ($ \hat{\boldsymbol{q}} = \{\hat{q}_{1}, \hat{q}_{2}, ... , \hat{q}_{7}\}$), velocity ($\dot{\hat{\boldsymbol{q}}} = \{\dot{\hat{q}}_{1}, \dot{\hat{q}}_{2}, ... , \dot{\hat{q}}_{7}\}$), and effort ($\hat{\boldsymbol{e}} = \{\hat{e}_{1}, \hat{e}_{2}, ... , \hat{e}_{7}\}$).
To jointly control the mobile base and the manipulator, the 6D pose of the end-effector relative to the mobile base is included ($\hat{X}^{mb}_{ee} \in \mathbb{R}^{6}$).
Finally, the mobile robot state is represented by the position and orientation relative to the initial pose of the door
($\hat{X}_{mb} \in \mathbb{R}^{3}$) as well as its velocity ($\dot{\hat{X}}_{mb}\in \mathbb{R}^{3}$).
\begin{equation} \label{state}
    s_t = \{\hat{\boldsymbol{q}}, \dot{\hat{\boldsymbol{q}}}, \hat{\boldsymbol{e}}, \hat{X}^{mb}_{ee}, \boldsymbol{S},
    \hat{X}_{mb}, \dot{\hat{X}}_{mb}\}_t \in \mathbb{R}^{53}
\end{equation}

The action space includes the joint velocity of the manipulator and the longitudinal and angular velocity of the mobile robot.
Since the manipulator has seven degrees of freedom and the mobile robot has two, there are a total of nine actions as follows:
\begin{equation} \label{action}
    a_t = \{\dot{\boldsymbol{q}}, \dot{X}_{mb}\}_t \in \mathbb{R}^{9}
\end{equation}

\subsubsection{Reward design}
\begin{figure}[b]%
\centerline{
    \subfloat[]{\includegraphics[width=0.3\columnwidth]{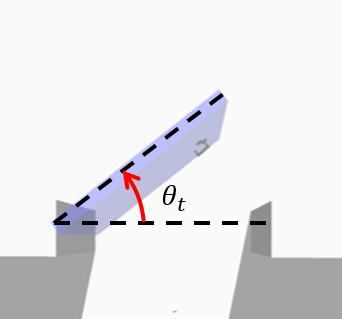} }%
    \subfloat[]{\includegraphics[width=0.3\columnwidth]{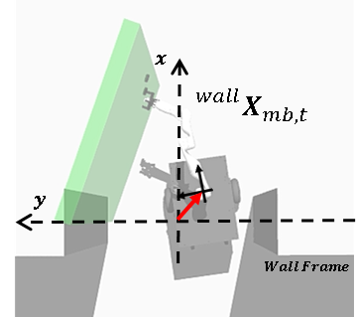} }%
    \subfloat[]{\includegraphics[width=0.3\columnwidth]{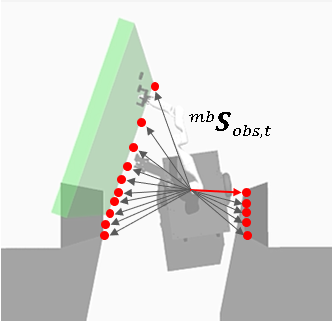} }%
}
\caption{Task reward definition. (a) door opening reward, (b) door passing reward, and (c) obstacle distance reward.}%
\label{task_reward}%
\end{figure}
Two main terms are used in reward ($r_t$) design to successfully open doors with stable motion: the task rewards $r_{mss,t}$, and the control penalties $r_{ctl,t}$.
\begin{equation} \label{r_total}
    r_t = r_{mss,t} + r_{ctl,t}
\end{equation}
For the $r_{mss,t}$, four task-related rewards are added with weights balancing: door open reward $r_{opn,t}$, door passing reward $r_{pss,t}$, obstacle distance reward $r_{obs,t}$, and collision penalty $r_{col,t}$.
\begin{equation}\label{r_mission}
    r_{mss,t} = w_{opn} r_{opn,t} + w_{pss} r_{pss,t}
    + w_{obs} r_{obs,t} + w_{col} r_{col,t}
\end{equation}
$r_{opn,t}$ is determined based on the extent to which the agent has opened the door in a single step.
It is proportional to the difference in the door angle between steps, which can be obtained from the ground truth in the simulation.
\begin{equation} \label{r_open}
    r_{opn,t} = \theta_t - \theta_{t-1}
\end{equation}
How much the robot succeeded in passing the door is reflected in $r_{pss,t}$, which includes the distance and angle of the robot with respect to the wall.
$r_{obs,t}$ represents the distance between the closest obstacle and the robot, which can be obtained from $\boldsymbol{S}$ in the state representation $s_t$.
\begin{equation} \label{r_obstacle}
    r_{obs,t} = - \underset{i}{min}\left(\boldsymbol{S}_{i,t}\right)
\end{equation}
\textbf{}
Finally, a collision penalty ($r_{col,t}$) of -1 was added whenever a self-collision or a collision between the robot and an obstacle occurred.

The control penalties include joint effort penalty $r_{eff,t}$, joint position penalty $r_{pos,t}$, and robot action scale penalty $r_{act,t}$.
\begin{equation} \label{r_control}
    r_{ctl,t} = w_{eff} r_{eff,t} + w_{pos} r_{pos,t} + w_{act} r_{act,t}
\end{equation}
Each penalty is scaled in $[-1, 0]$, and is expressed in polynomial as shown in equation (\ref{r_effort}) that it can increase rapidly around the limit value.
$r_{pos,t}$ and $r_{eff,t}$ include joint position penalty and effort penalty for the seven joints of the manipulator.
Meanwhile, $r_{act,t}$ includes both the manipulator and mobile robot actions to prevent unnecessary actions and to smooth out the movement.
\begin{equation} \label{r_effort}
    r_{eff,t} = -\sum_{j=1}^{7} w_{e,j} \left(e_{j,t}\right)^2
\end{equation}

\subsubsection{Algorithm and architecture}
We use the soft actor-critic (SAC) algorithm \cite{SAC} with the double Q-learning scheme due to its sample efficiency and stability concerning hyperparameter tuning, which has been proven in the field of various robotic applications, including autonomous driving \cite{fuchs2020super}, \cite{seong2021learning} and manipulator control \cite{yang2019control}, \cite{prianto2020path}.
 Algorithm \ref{algorithm:SAC} summarizes our overall training process with a hyperparameter configuration in Table \ref{table:RL_hparameters}.
 Following the approach in \cite{haarnoja2018softapplication}, \cite{fujimoto2018addressing}, we implement an actor-critic framework that consists of a policy network $\pi_{\phi}$ and two Q-value networks $Q_{\theta_1}$ and $Q_{\theta_2}$ for the double Q-learning approach.
 In an episode loop, we first obtain state information $s_t$ from the manipulator and mobile robot. Our policy then infers an action $a_t$ and interacts with the environment to get state transition data $({s}_{t}, {a}_{t}, r({s}_{t}, {a}_{t}), {s}_{t+1})$, which is saved in a replay buffer $D$.
During the update phase, the SAC algorithm optimizes each of the Q-value networks by minimizing the Bellman error of the soft Q-function ($J_Q(\theta_i)$). Then the algorithm updates the policy by minimizing the objective function $J_{\pi}(\phi)$ composed of the expected return and entropy bonus as below:
 \begin{equation}
    \label{eq:objective_function}
J_{\pi}(\phi) = - \mathbb{E}_{s_t \sim D} [\mathbb{E}_{a_t \sim \pi_{\phi}} [\min_{i=1,2}Q_{\theta_i}(s_t, a_t) + \alpha H(\pi_{\phi})]],
\end{equation}
where $s_t$ and $a_t$ are sampled to compute the Q-values from the buffer $D$ and policy $\pi_{\phi}$, respectively, and the minimum of the two Q-values is chosen to mitigate value overestimation. The entropy term $H(\pi_{\phi})$, with an automatically tunable coefficient $\alpha$, acts as a regularizer to maintain proper uncertainty in the policy model. By balancing the clipped Q-value and entropy of the policy, the SAC algorithm achieves a trade-off between maximizing the expected return and exploring near-optimal policies.

\begin{algorithm}[h!]
\caption{SAC with the double Q-learning scheme.}
\label{algorithm:SAC}
\textbf{Input: } $\phi, \theta_{1}, \theta_{2} $ \\
\textbf{Results: } \text{Optimized parameters } $\phi, \theta_{1}, \theta_{2} $

\begin{algorithmic}[1]
\State Initialize parameters $\phi, \theta_{1}, \theta_{2}$ and dataset $D \leftarrow \emptyset$

\For{each episode}
    \While{not done}
        \State $ {a}_{t} \sim \pi_{\phi}({a}_{t} | {s}_{t}) $
        \State $ {s}_{t+1} \sim p({s}_{t+1} | {s}_{t}, {a}_{t})  $
        \State $ D \leftarrow D \cup \{({s}_{t}, {a}_{t}, r({s}_{t}, {a}_{t}), {s}_{t+1})\} $
    \EndWhile
    \For{each gradient step}
        \State $ \theta_{i} \leftarrow \theta_{i} - \lambda_{Q} \nabla_{\theta_{i}} J_{Q}(\theta_{i}) $ for $i \in \{1,2\} $
        \State $ \phi \leftarrow \phi - \lambda_{\pi} \nabla_{\phi} J_{\pi}(\phi) $
        \State $ \alpha \leftarrow \alpha - \lambda \nabla_{\alpha} J(\alpha) $ 
    \EndFor
\EndFor

\end{algorithmic}
\end{algorithm}

\begin{figure}[t]%
    \centerline{\includegraphics[width=0.9\columnwidth]{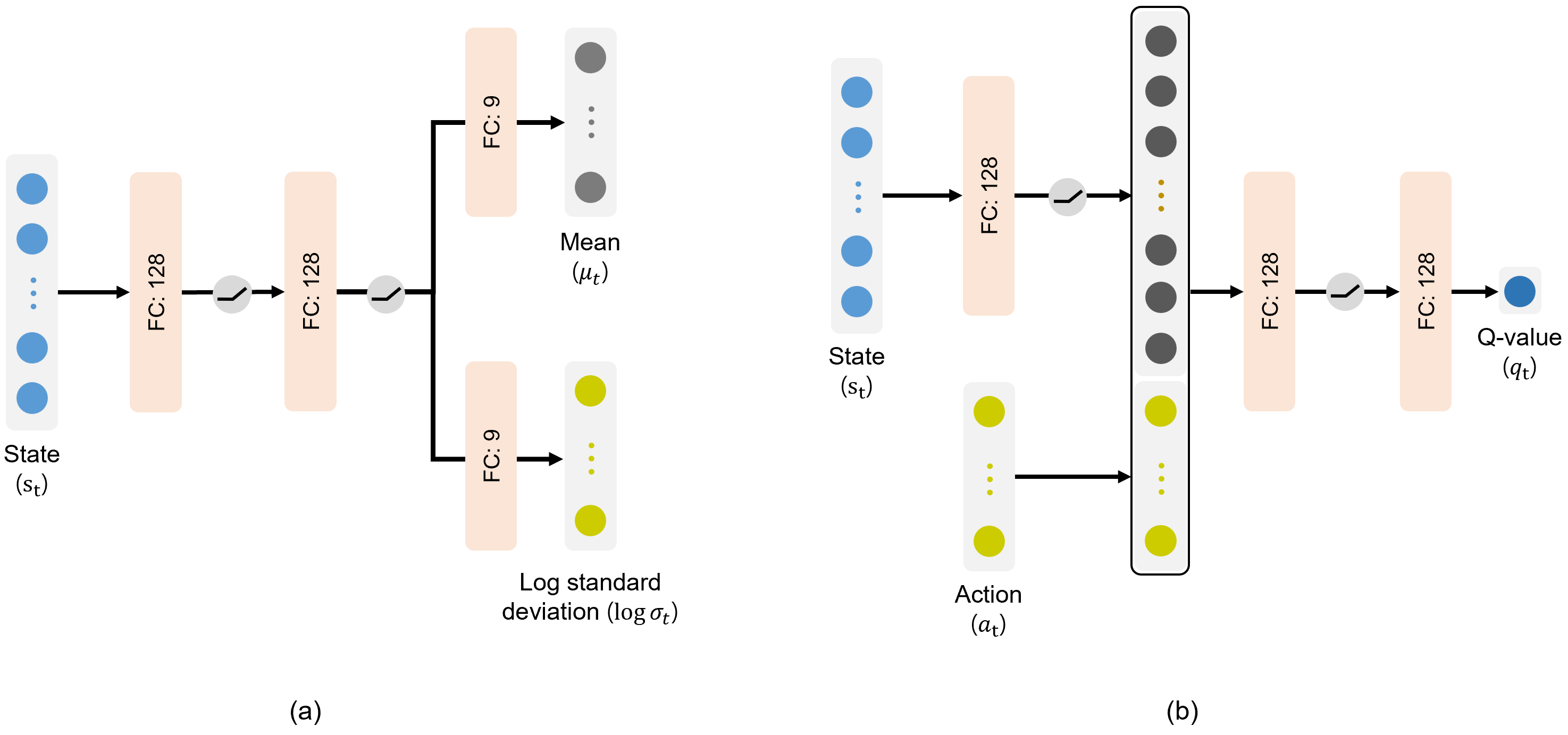}}
\caption[]{Actor-critic network architecture. (a) policy network, and (b) Q-value network.}%
\label{fig:network}%
\end{figure}

Fig. \ref{fig:network} illustrates the designed policy and Q-value networks. A squashed Gaussian policy is used to infer actions in a continuous space \cite{haarnoja2018softapplication}.
To model the policy network, we use two sequential fully connected (FC) layers with ReLU nonlinearity, each consisting of 128 neurons, and two branched FC layers with 9 neurons that encode the state $s_t$ into the mean $\mu_t$ and log standard deviation $log\sigma_t$ of the action $a_t \in \mathbb{R}^{9}$.
The policy samples stochastic actions by a Gaussian distribution in the training phase and directly outputs deterministic actions with the mean in the evaluation phase.
Since the sampled or deterministic action is not bounded in a finite range, a nonlinear squashing function, hyperbolic tangent, is used to ensure the action is in $[-1,1]$ as follows:
\begin{equation}
    \label{eq:squashed_gaussian}
a_t = \tanh(\mu_{t} + \sigma_{t} \odot \epsilon), \quad \epsilon \sim N(0, I),
\end{equation}
where $\odot$ is the element-wise product.
Unlike the policy network, the Q-value network takes both the state $s_t$ and action $a_t$ as input data to evaluate the policy.
We design the network with a single FC layer consisting of 128 neurons and ReLU activation to encode state input.
The encoded feature is then concatenated with the action and processed by two additional hidden layers to compute the Q-value of the given state and action ($Q(s_t, a_t)$).

\begin{table}[t]
\caption[Hyper-parameter configuration for training]{Hyper-parameter configuration for training.}
\label{table:RL_hparameters}
\begin{center}
\begin{tabular} {ccccccccccc}
\hline\hline
& Hyper-parameter& value\\
\hline
& Actor's learning rate ($\lambda_Q$) & 0.0003\\
& Critic's learning rate ($\lambda_{\pi}$) & 0.0003\\
& Alpha learning rate ($\lambda_{\alpha}$)& 0.0003\\
& Batch size & 1024\\
& Step time & 0.2s\\
& Max step per episode & 100\\
& Training iteration & 10 episodes\\
\hline\hline
\end{tabular}
\end{center}
\end{table}

\section{Experiment setup}
\label{sec:experiments}
\subsection{Mobile manipulator hardware}
\label{sec:robot_system}
\begin{figure}[h!]
    \centerline{\includegraphics[width=0.8\columnwidth]{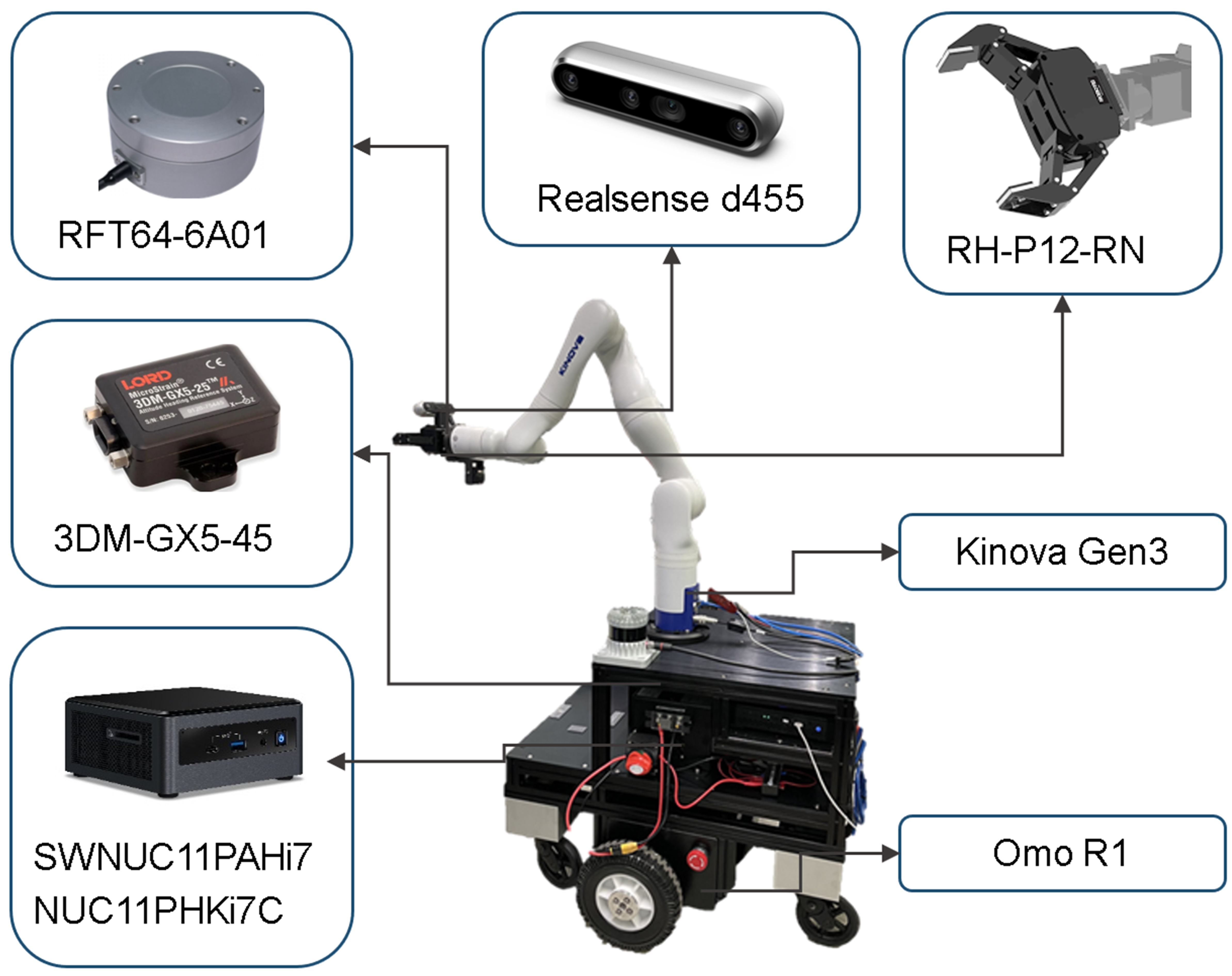}}
    \caption{Hardware configuration and sensors used for the mobile manipulator.}
    \label{fig:hardware}
\end{figure}
The hardware configuration of our mobile manipulator system is depicted in Fig. \ref{fig:hardware}.
The mobile manipulator consists of the Kinova Gen3 manipulator with 7 degrees of freedom and the Omo R1 mobile base platform equipped with differential wheels.
Robotis two-finger gripper (PH-P12-RN) is used as the robot's hand, which is capable of easily grasping various types of door handles including levers, door knobs, and bar handles.
The gripper has a payload capacity of 5kg, a gripping force of 170N, and weighs only 500g, making it suitable for the mobile manipulator due to its lightness compared to other finger designs.
A force/torque sensor (RFT64-6A01) for external force measurement and an Intel RealSense D455 depth camera for object recognition is attached between the manipulator and the gripper.
The robot utilizes two Intel NUCs as onboard computers, with one of them, the NUC11PHKi7C, running the deep learning algorithms.

\subsection{Training environment}
\begin{figure}[h!]
    \centerline{\includegraphics[width=0.8\columnwidth]{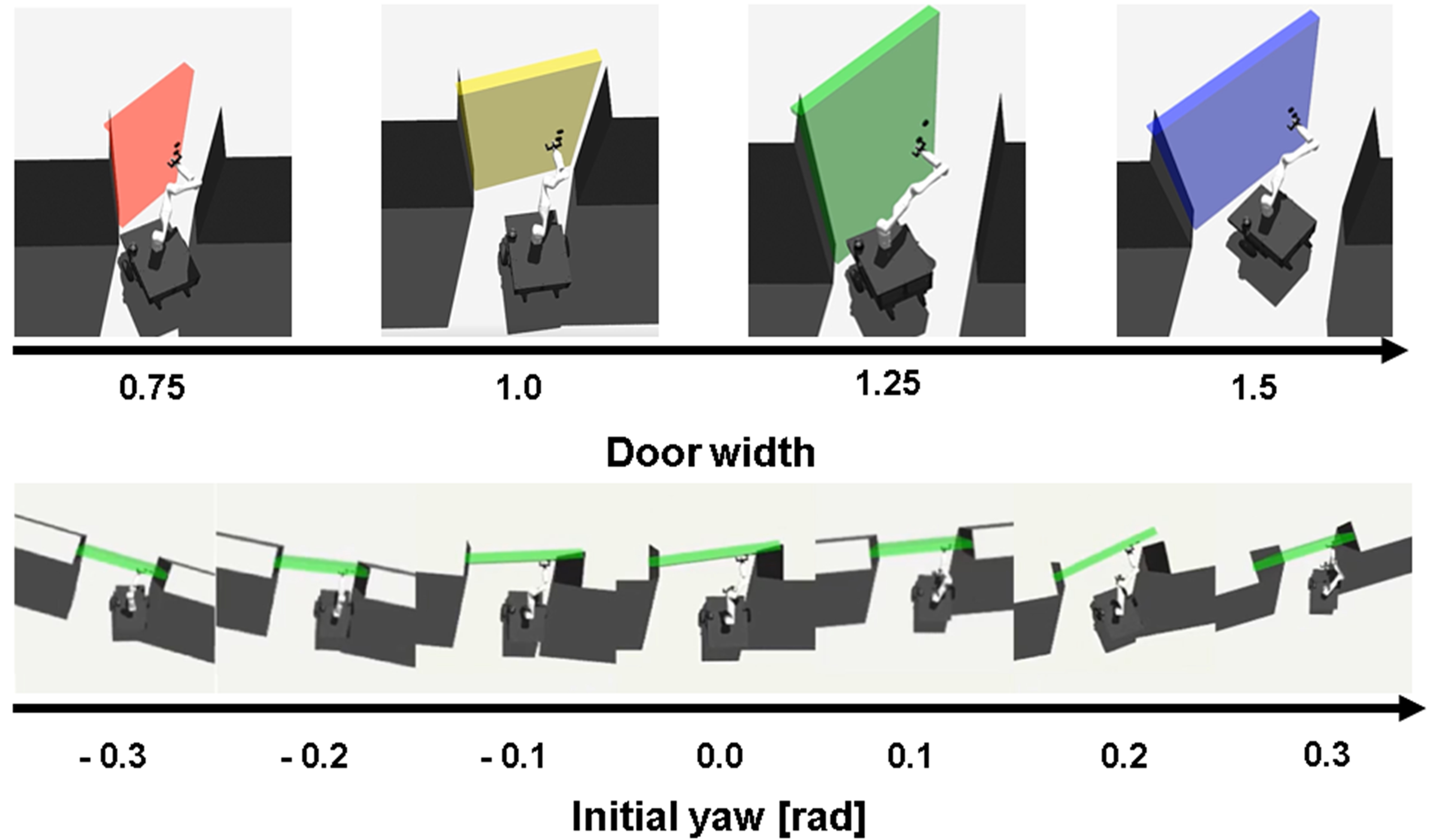}}
    \caption{Training scenarios with four different door widths and seven initial poses.} \label{random_scenarios}
\end{figure}
We configured the training environment using the GAZEBO simulator with Robotic Operation System (ROS) for communication between the robot and the simulated environment.
The state was then preprocessed and fed to the policy model, which inferred and transferred actions through ROS messages.
All computed data of the state, action, and reward pairs were collected in a replay buffer for training. We randomized the width of doors (0.75 to 1.5 m) and the initial pose angle of the base platform (-0.3 to 0.3) in the environment at the beginning of each episode, to enhance the robustness of our policy model as in Fig. \ref{random_scenarios}.
We also added observation noise to the LiDAR data as Gaussian distribution with 0.04 standard deviation for a more realistic configuration.

\section{Results}
\label{sec:results}
\subsection{Handle pose estimation}
\begin{figure}[t]%
\centerline{
\subfloat[]{\includegraphics[width=0.55\columnwidth]{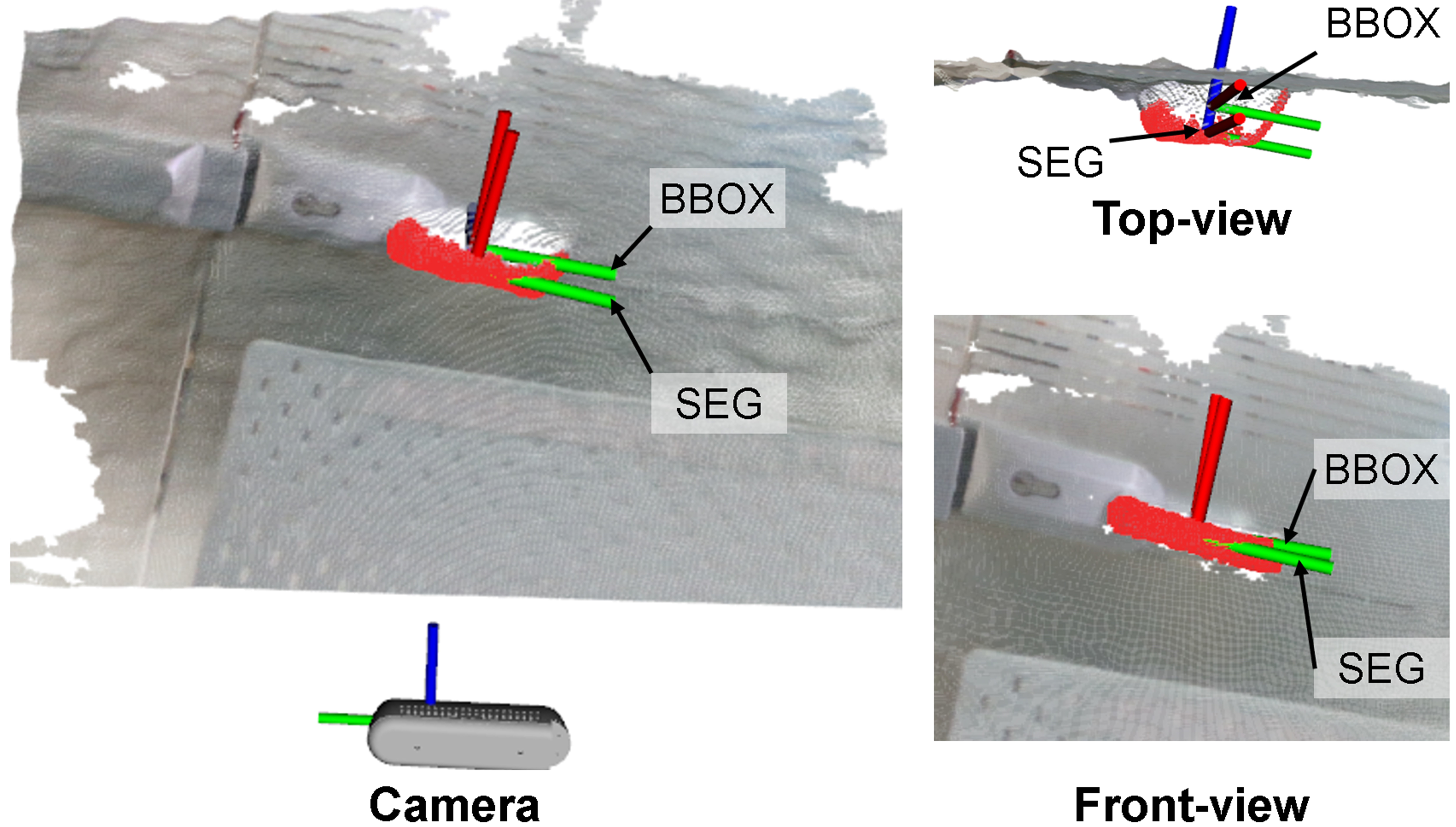}}
}
\centerline{
\subfloat[]{\includegraphics[width=0.23\columnwidth]{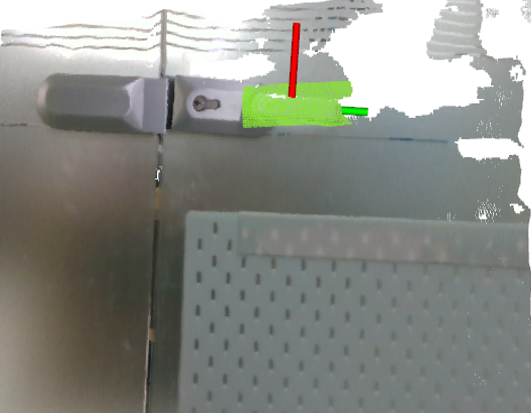}} \hfill
\subfloat[]{\includegraphics[width=0.23\columnwidth]{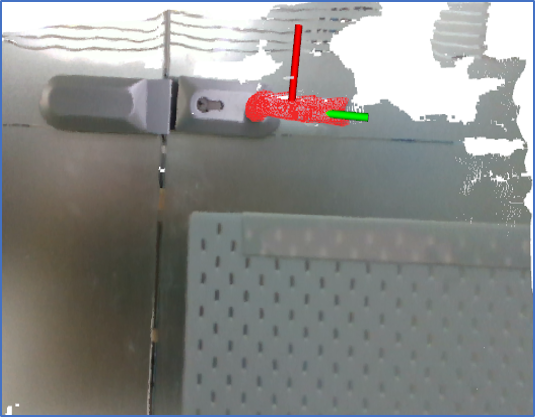}} \hfill
\subfloat[]{\includegraphics[width=0.23\columnwidth]{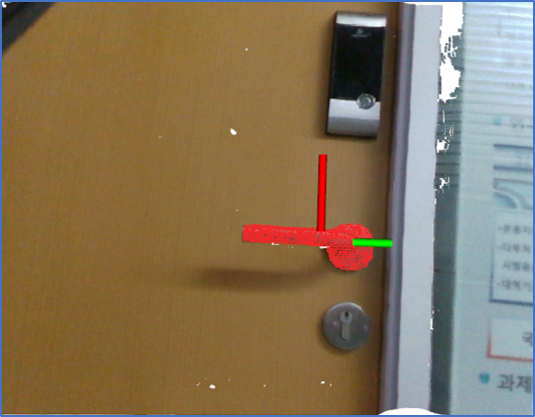}} \hfill
\subfloat[]{\includegraphics[width=0.23\columnwidth]{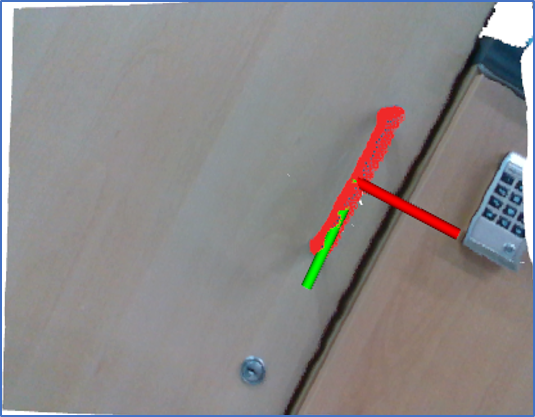}}
}
\caption{Detected handle point cloud and its pose estimation. (a) Pose estimation using the bounding box-based (BBOX) and semantic segmentation-based (SEG) algorithms. (b) Detected handle point cloud (green) and estimated handle pose using the bounding box. (c) Detected handle point cloud (red) and estimated handle pose using our semantic segmentation approach. (d)-(e) Handle detection and pose estimation for different handle shapes and types with the semantic segmentation-based algorithm. The estimated poses shown in the figure are in the camera frame.} \label{fig:handle_type1_bbx_and_seg}
\end{figure}

The handle pose estimation algorithm was tested on handles with different shapes.
Fig. \ref{fig:handle_type1_bbx_and_seg} (a) and (c-e) demonstrate the successful extraction of the target point cloud (highlighted in red) from the depth image.
As the image segmentation which is used as the mask to filter the target handle, was able to detect object regardless of its shape and size, the detection algorithm was also able to detect various handles.
Moreover, the algorithm showed its effectiveness in estimating the handle pose across different camera orientations.
It successfully estimated the pose even when the camera pose was tilted, in contrast to the algorithm that relied on using a bounding box as a mask (b).
Compared to the bounding box method, which could not accurately separate the background form the target, the segmentation-based approach exhibited higher accuracy in depth (0.027 $m$) and orientation estimation.
The algorithm also successfully detected the handle on both transparent and non-transparent doors.

\subsection{Training in virtual environment}

\begin{figure}[h!]
    \captionsetup[subfigure]{aboveskip=-9pt,belowskip=-9pt}
    \centering
        \begin{subfigure}[b]{0.35\columnwidth} {\includegraphics[width=\columnwidth]{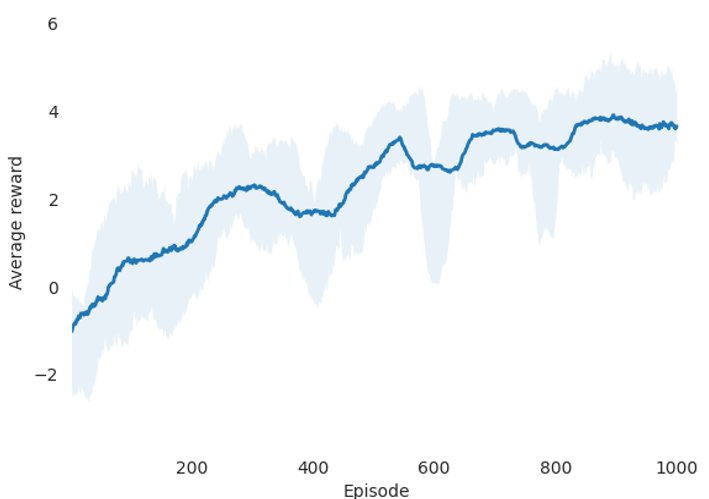} } \caption{} \end{subfigure}
    \centering
        \begin{subfigure}[b]{0.35\columnwidth} {\includegraphics[width=\columnwidth]{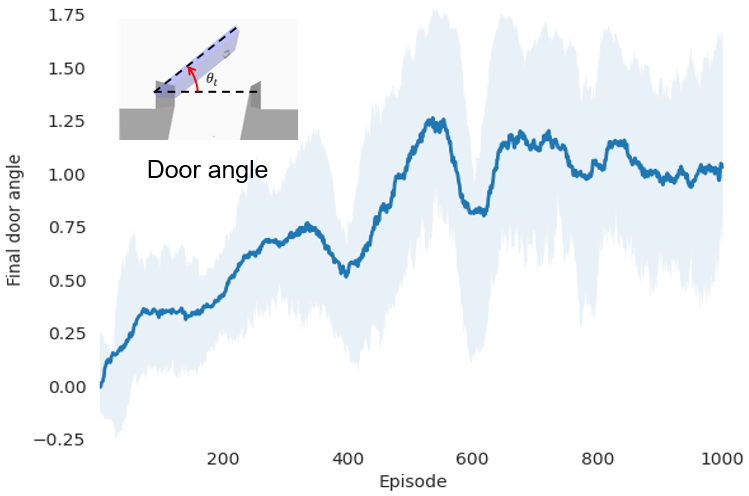} } \caption{} \end{subfigure}
\caption[]{Training results. (a) average reward, and (b) the final door angle of each episode.}%
\label{fig:training_result}%
\end{figure}
Training the agent for door opening with RL in simulation went through 1000 episodes, with a maximum of 100 steps per episode.
Episodes were terminated either when the robot opened the door up to a maximum angle of $\pi/2$ or when a collision occurred.
Fig. \ref{fig:training_result} presents the average training reward (a) and the final door angle achieved in each episode (b).
The robot was trained only for the CCW push-open door as the model could not converge when given scenarios with diverse door opening directions.

\begin{table*}[ht]
\centering
\caption{Quantitative evaluation of door opening in simulated scenarios. Door widths: Wide (1.25 m), Mid (1.0 m), Narrow (0.75 m). SRate and T represent the success rate (\%) and the time (s) taken to complete the door opening process accordingly.}
\label{tab:rl_result}
\scriptsize
\begin{tabular*}{\textwidth}{@{\extracolsep{\fill}}c*{8}{S}}
\toprule
& \multicolumn{8}{c}{Door width} \\
\cmidrule{2-9}
& \multicolumn{2}{c}{Wide} & \multicolumn{2}{c}{Mid} & \multicolumn{2}{c}{Narrow} & \multicolumn{2}{c}{0.9\, \si{m} (unseen)} \\
\cmidrule(lr){2-3}\cmidrule(lr){4-5}\cmidrule(lr){6-7}\cmidrule(lr){8-9}
{Initial yaw} & {\makecell{SRate $\uparrow$}} & {\makecell{T $\downarrow$}} &{\makecell{SRate $\uparrow$}} &{\makecell{T $\downarrow$}} &{\makecell{SRate $\uparrow$}} &{\makecell{T $\downarrow$}} &{\makecell{SRate $\uparrow$}} &{\makecell{T $\downarrow$}} \\
\cmidrule{1-9}
Left &1.0 &8.45 &1.0 &8.18 &1.0 &11.18 &1.0 &7.80 \\
Center &1.0 &11.66 &1.0 &7.66 &1.0 &6.28 &1.0 &10.54 \\
Right &1.0 &9.08 &1.0 &5.38 &0.50 &14.62 &0.77 &13.36 \\
\textbf{Average} &\textbf{1.0} &\textbf{9.73} &\textbf{1.0} &\textbf{7.07} &\textbf{0.83} &\textbf{10.69} &\textbf{0.92} &\textbf{10.57} \\
\bottomrule
\end{tabular*}
\end{table*}

To evaluate the proposed learning-based policy, we conducted tests on twelve scenarios, nine of which were training scenarios, and three were unseen scenarios.
The scenarios included four types of door widths ($0.75, 0.9, 1.0,$ and $1.25 m$), with one unseen width of $0.9 m$.
The initial poses of the robot relative to the target door were chosen among three different yaw: Left ($-0.3$), Center ($0.0$), and Right ($0.3$).
During the evaluation, an episode was considered successful if the robot opened the door wider than its width of $0.66 m$.
Conversely, an episode was considered a failure if the robot reached the time limit or experienced a collision.

The evaluation results in Table \ref{tab:rl_result} show that the robot successfully opened the door in most cases.
The robot did not achieve a 1.0 success rate only for the narrowest door ($0.75 m$) and the door with unseen width ($0.9 m$) that was tilted to the right side.
In particular, it showed the lowest success rate of 0.5 with the narrow door tilted to the right, as the robot kept having collisions with the door and the wall due to the initial pose of its elbow joint being close to the right side of the wall.
However, the robot achieved a 1.0 success rate for non-tilted and left-tilted 0.9 $m$ width doors, which shows that the robot is able to open doors that were not part of its training data.

The average opening time was approximately 10 seconds for wide, narrow, and 0.9 $m$ doors, while the time for the 1.0 $m$ door was 7.07 seconds.
The tilted-narrow door took the longest time to open, with an average of 14.62 seconds, while the non-tilted narrow door was the quickest at 6.27 seconds.
This difference was due to the initial pose of the manipulator joints, which caused the robot to move its joints further from the wall to avoid collisions during the door opening motion.
The maximum time spent opening the door was 14.8 seconds, which occurred with the tilted-narrow and 0.9 $m$ doors, and the minimum time spent was 4.8 seconds for the non-tilted narrow door.

\subsection{Door opening in the real-world}
\label{sec:real-world-imp}
\begin{figure*}[]
\centering
\subfloat{\includegraphics[width=1.0\textwidth]{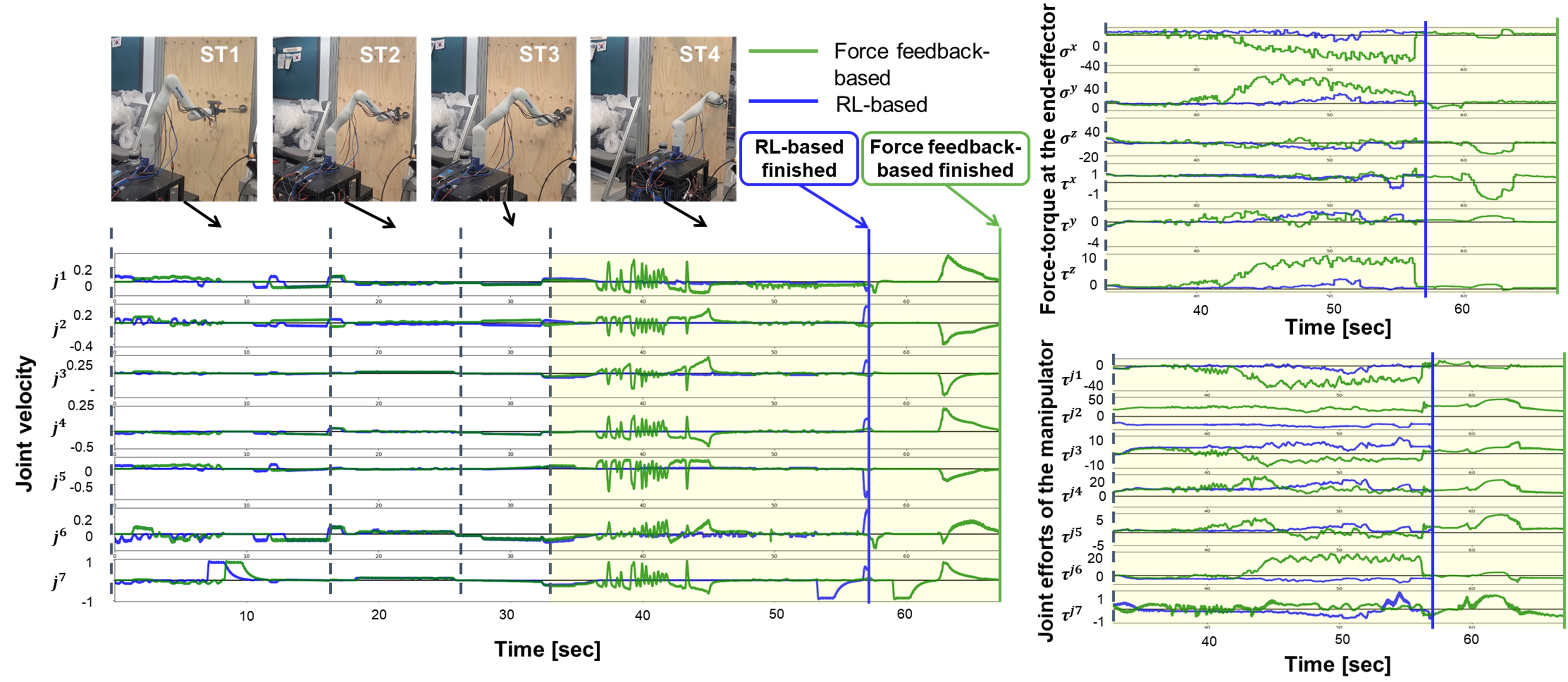}}%
\caption{Real-world demonstration and performance of adaptive position-force-based and RL-based door opening motion. Yellow areas represent the result of subtask 4 of each system. The green lines are the results of the adaptive position-force control-based system, and the blue lines represent the RL-based system. Images of the robot in each subtask phase are at the top left figure.} \label{fig:performance_result}
\end{figure*}
The system's performance was evaluated based on the stability of joint velocities, force/torque applied at the end-effector, and efforts on the manipulator's joints.
Fig. \ref{fig:performance_result} illustrates the real-world test results for both the adaptive position-force-based and RL-based door opening motion.
The test was conducted on a Push-CCW type door with a width of $0.9 m$ and a lever type (type1) handle, where $\psi^x = 1$ and $\psi^y = 1$.
Since the RL algorithm was only trained for the door opening and passing motion, the other phases (ST1-ST3) were executed in the same way as the adaptive position-force-based system.

The average force and torque applied at the end-effector were $8.82\ N$ and $1.29\ N\cdot m$ for the adaptive position-force control and $5.75\ N$ and $0.72\ N\cdot m$ for the RL-based control.
When using the adaptive position-force control, robot did not generate an accurate trajectory due to lack of information, but rather moved roughly towards the opening direction of the door and adjusted to the external forces by feedback control.
This caused a large force at the end-effector, as the robot did not estimate the force it would cause by moving in such constrained environment.
However, RL agent was able to reduce the maximum force by 3.27 times and the maximum torque by 1.29 times, which indicates that the robot was able to learn actions that could reduce the external force and torque.
The average effort on the joints was $6.73\ N\cdot m$ and $5.41\ N\cdot m$ for the adaptive position-force-based system and the RL-based system, respectively, which also shows that the RL agent effectively reduced torques applied on each joints.

The average joint acceleration of the adaptive position-force-based system was $0.042 sec^{-2}$, while it was $0.023 sec^{-2}$ for the RL-based system, which was almost half that of the adaptive position-force-based system.
The gap was pronounced at the end-effector joint (joint 7), where the average angular acceleration was $0.073 sec^{-2}$ for the adaptive position-force-based and $0.040 sec^{-2}$ for the RL-based method.
The adaptive position-force-based method showed high joint acceleration before the robot identified the horizontal opening direction ($\psi^{y}$) of the door, as the robot reacted more sensitively to external forces.
However, the stability of the robot's actions was quite similar afterward, indicating that both agents were able to open the door smoothly once they knew how to open it.
Additionally, the RL-based system completed the task more quickly.

\subsection{Versatile door opening with adaptive position-force control}

\begin{figure}[h!]%
    \centerline{\includegraphics[width=\columnwidth]{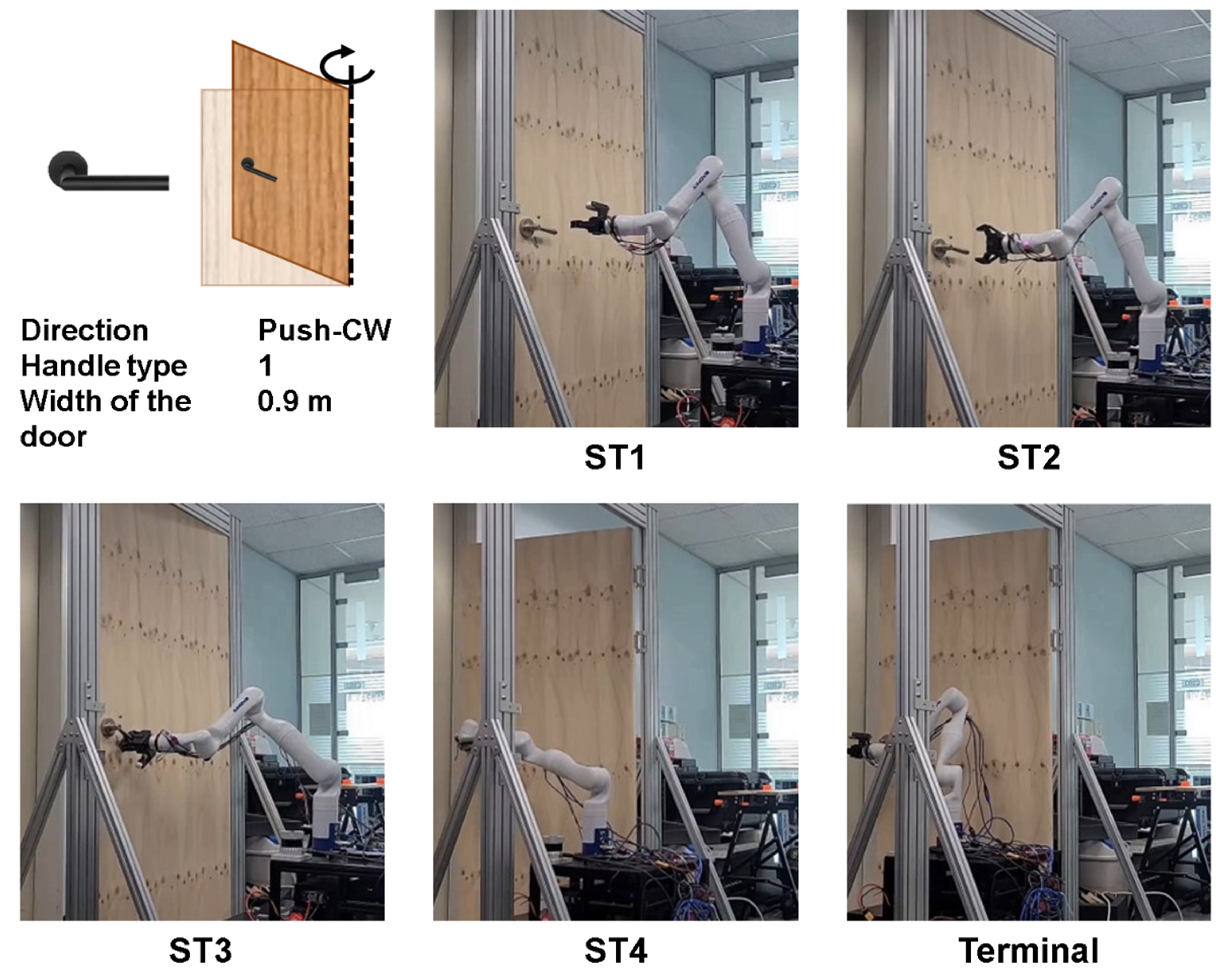}}
    \caption{The opening process for a Push-CW type door with a type1 handle starts with the robot detecting and classifying the handle in ST1, and preparing to unlatch it in ST2. Once the handle is unlatched, the robot determines the door's opening direction in ST3. Since the door is a Push-CW type, the robot moves its end-effector in the front-left direction while compensating for external forces generated during door the opening motion using adaptive position-force control in ST4. When the door is fully open, the robot returns to its manipulator pose and initiates autonomous driving mode as the terminal state.}
    \label{fig:demo_1}
\end{figure}
\begin{figure}[h!]%
    \centering {\includegraphics[width=\columnwidth]{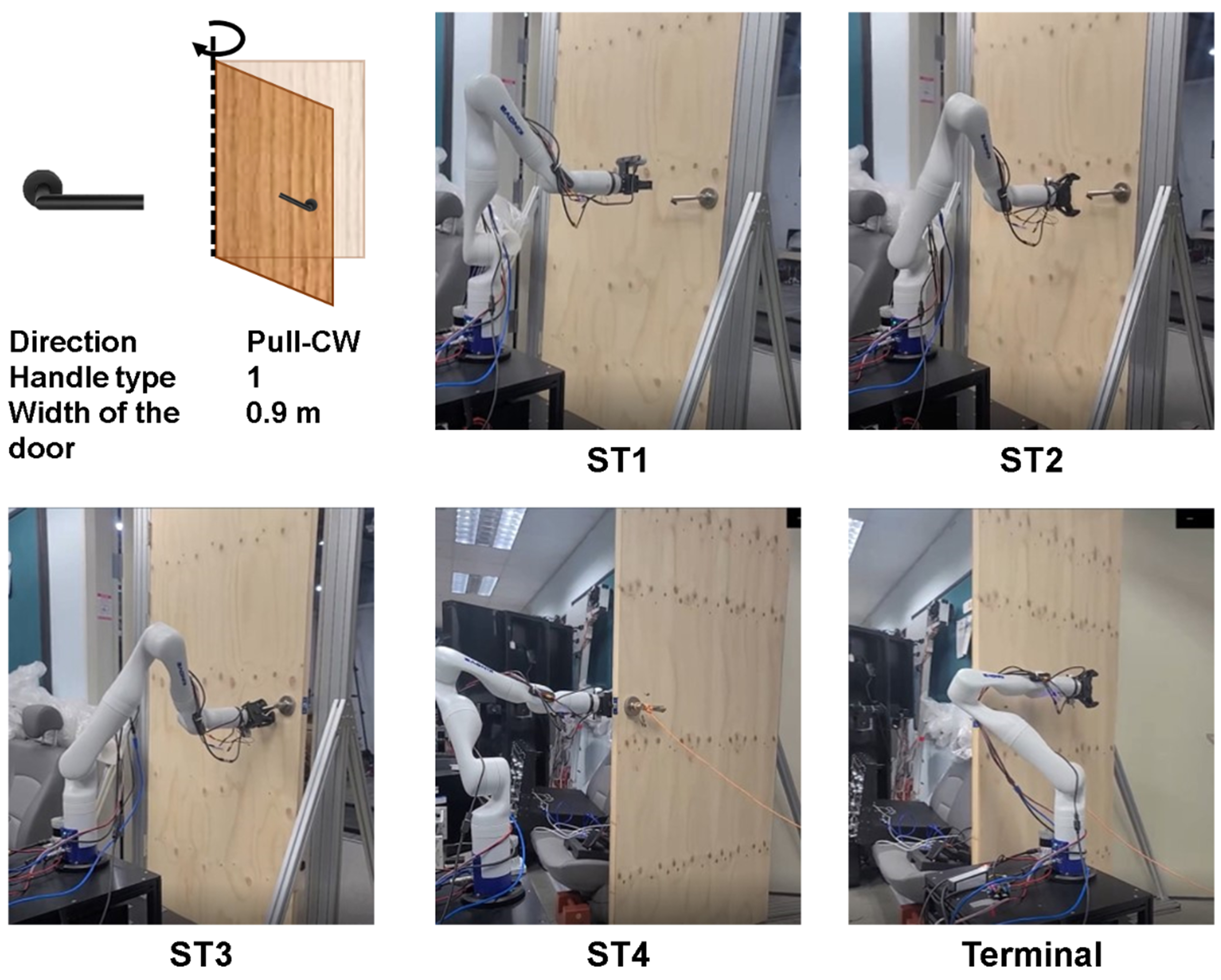} }
\caption{Opening Pull-CW type door with type1 handle. The robot estimates the pose of the handle at ST1 and grabs and unlatchs the handle according to its type in ST2. The robot identifies the door is a Pull-CW type in ST3. While opening the door, the mobile base moves backward to avoid collision with the door and generate the manipulator workspace in ST4. When the door is fully opened, the mobile robot moves forward to block the door from closing and release the end-effector and initialize the manipulator pose.} 
\label{fig:demo_2}%
\end{figure}
\begin{figure}[h!]%
    \centering
        {\includegraphics[width=\columnwidth]{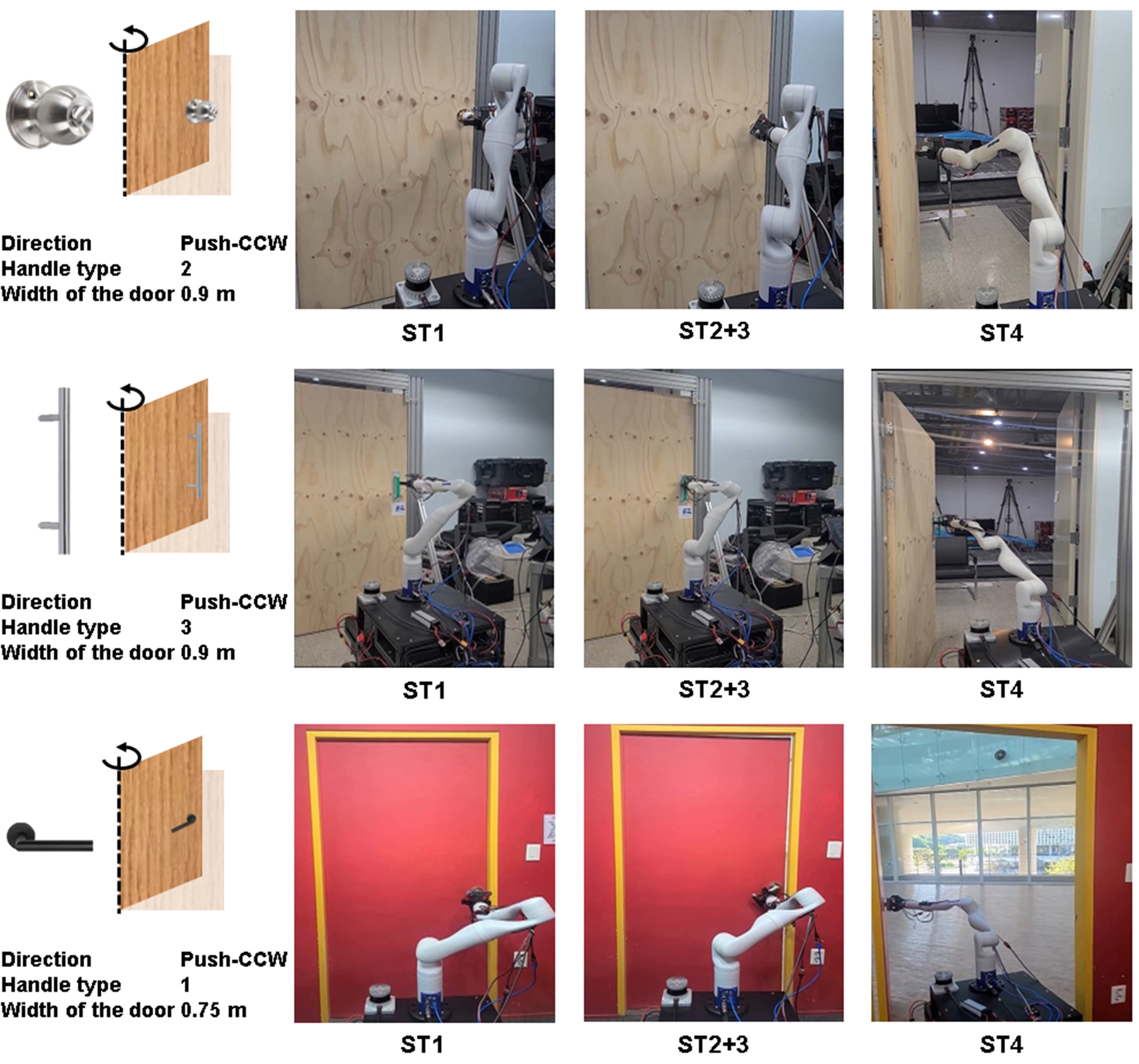} }
\caption{Opening Push-CCW type door with type1, type2, and type3 handle. If the target handle is type2, the robot grasps the handle and turns it until $\sigma_{ee}$ reaches $\Omega_{ee}$ and $\Lambda_{ee}$. However, when the handle is type3, the robot grasps the handle and directly moves on to ST3 without unlatching motion.
The robot successfully passes through the door with different widths and different shapes of handles (type1) in the last row of the figure using adaptive motion.}
\label{fig:demo_3}%
\end{figure}

The adaptive position-force control-based system was evaluated with doors of various kinematics, such as different widths, opening directions $\psi^x$, $\psi^y$, and handle types, as illustrated in Fig. \ref{fig:demo_1}-\ref{fig:demo_3}.
In Fig. \ref{fig:demo_1}, the robot opened a Push-CW type door ($\psi^x = 1$, $\psi^y = -1$) with a width of 0.9m.
In ST1, the robot approached the door from 2.5m away while simultaneously estimating the 6D pose of the handle and aligning with it.
After aligning with the handle, the robot reached and unlatched the handle based on its type (type1) using the end-effector in ST2.
Our method properly unlatched the handle until the force at the end-effector detects the unlock based on the real-time force/torque data.
In ST3, the robot estimated the kinematics of the door, including its opening direction with force measurement.
Finally, the robot opened the door based on the estimated door kinematics (ST4) until the door was fully opened, allowing the robot to pass through in the terminal stage.

In Fig. \ref{fig:demo_2}, our robot succeeded in opening a more challenging scenario of the Pull-CW type door ($\psi^x = -1$, $\psi^y = 1$).
Similar to the scenario of the Push-CW type door, the robot drew close to the door and unlocked the handle (type1) during ST1-ST2 while estimating the door kinematics in the process of ST3.
Since our method determined it as a pulling door, the end-effector of the manipulator performed pulling actuation while the base robot reactively maneuvered backward based on the real-time workspace checking algorithm.
After opening the door so that our mobile base robot had enough drivable region, the robot drove forward to go through the door, and the end-effector returned to its initial pose.

The door opening system was also tested on type2 and type3 handles in Fig. \ref{fig:demo_3}.
The procedure was the same as the door with type1 handle for both types, except for ST2 in the type3 handle as it does not need to be unlocked.
To test the adaptiveness of the door opening motion, doors with different widths were tested in the bottom figures of Fig. \ref{fig:demo_3}.
The robot successfully opened the type1 handle with different shapes and sizes and was able to adaptively open the narrow door with adaptive position-force control.

\section{Conclusion}
\label{sec:conclusion}
In this paper, we present a mobile manipulator system capable of full sequence opening and passing doors without prior knowledge or human guidance.
The task was decomposed into four subtasks, and relevant information such as the pose, type of the handle, and the opening direction of the door was identified online during each procedure.
Also, we propose two most popular approaches that enable the mobile manipulator to open doors: an adaptive position-force-based control method and a reinforcement learning-based method.

The adaptive position-force controller enabled the robot to open doors in four different directions and widths, demonstrating its flexibility in terms of different door trajectories without prior knowledge.
However, it exhibited chattering in joint velocity and high torque on each joint.
However, it exhibited chattering in joint velocity and high torque on each joint due to its reliance solely on the end-effector force/torque state and sensitivity to sensor noise. 
On the other hand, the RL-based control system considered not only forces but also the position, velocity, and surrounding obstacles, enabling it to perform well-defined tasks.
It achieved stable behavior while opening doors with varying widths, initial poses, and previously unseen kinematics. 
However, as the task scope expands, it has been observed that RL failed to train the policy, revealing limitations in terms of scalability to diverse tasks.

\section*{Acknowledgement}
This research was supported by the Challengeable Future Defense Technology Research and Development Program through the Agency For Defense Development(ADD) funded by the Defense Acquisition Program Administration(DAPA) in 2023(No.915027201).




\begin{thebibliography}{00}
\bibitem{lee2021assistive}Lee, D., Kang, G., Kim, B. \& Shim, D. Assistive Delivery Robot Application for Real-World Postal Services. {\em IEEE Access}. \textbf{9} pp. 141981-141998 (2021)
\bibitem{srinivas2022autonomous}Srinivas, S., Ramachandiran, S. \& Rajendran, S. Autonomous robot-driven deliveries: A review of recent developments and future directions. {\em Transportation Research Part E: Logistics And Transportation Review}. \textbf{165} pp. 102834 (2022)
\bibitem{tanner2001advanced}Tanner, H., Kyriakopoulos, K. \& Krikelis, N. Advanced agricultural robots: kinematics and dynamics of multiple mobile manipulators handling non-rigid material. {\em Computers And Electronics In Agriculture}. \textbf{31}, 91-105 (2001)
\bibitem{iriondo2019pick}Iriondo, A., Lazkano, E., Susperregi, L., Urain, J., Fernandez, A. \& Molina, J. Pick and place operations in logistics using a mobile manipulator controlled with deep reinforcement learning. {\em Applied Sciences}. \textbf{9}, 348 (2019)
\bibitem{yang2019collaborative}Yang, M., Yang, E., Zante, R., Post, M. \& Liu, X. Collaborative mobile industrial manipulator: a review of system architecture and applications. {\em 2019 25th International Conference On Automation And Computing (ICAC)}. pp. 1-6 (2019)
\bibitem{peterson2000high}Peterson, L., Austin, D. \& Kragic, D. High-level control of a mobile manipulator for door opening. {\em Proceedings. 2000 IEEE/RSJ International Conference On Intelligent Robots And Systems (IROS 2000)(Cat. No. 00CH37113)}. \textbf{3} pp. 2333-2338 (2000)
\bibitem{ahmad2012multiple}Ahmad, S., Zhang, H. \& Liu, G. Multiple working mode control of door opening with a mobile modular and reconfigurable robot. {\em IEEE/ASME Transactions On Mechatronics}. \textbf{18}, 833-844 (2012)
\bibitem{ma2018optimal}Ma, C., Gao, H., Ding, L., Tao, J., Xia, K., Yu, H. \& Deng, Z. Optimal energy consumption for mobile manipulators executing door opening task. {\em Mathematical Problems In Engineering}. \textbf{2018} (2018)
\bibitem{stuede2019door}Stuede, M., Nuelle, K., Tappe, S. \& Ortmaier, T. Door opening and traversal with an industrial cartesian impedance controlled mobile robot. {\em 2019 International Conference On Robotics And Automation (ICRA)}. pp. 966-972 (2019)
\bibitem{arduengo2021robust}Arduengo, M., Torras, C. \& Sentis, L. Robust and adaptive door operation with a mobile robot. {\em Intelligent Service Robotics}. \textbf{14}, 409-425 (2021)
\bibitem{minniti2019whole}Minniti, M., Farshidian, F., Grandia, R. \& Hutter, M. Whole-body mpc for a dynamically stable mobile manipulator. {\em IEEE Robotics And Automation Letters}. \textbf{4}, 3687-3694 (2019)
\bibitem{luo2019reinforcement}Luo, J., Solowjow, E., Wen, C., Ojea, J., Agogino, A., Tamar, A. \& Abbeel, P. Reinforcement learning on variable impedance controller for high-precision robotic assembly. {\em 2019 International Conference On Robotics And Automation (ICRA)}. pp. 3080-3087 (2019)
\bibitem{johannink2019residual}Johannink, T., Bahl, S., Nair, A., Luo, J., Kumar, A., Loskyll, M., Ojea, J., Solowjow, E. \& Levine, S. Residual reinforcement learning for robot control. {\em 2019 International Conference On Robotics And Automation (ICRA)}. pp. 6023-6029 (2019)
\bibitem{beltran2020variable}Beltran-Hernandez, C., Petit, D., Ramirez-Alpizar, I. \& Harada, K. Variable compliance control for robotic peg-in-hole assembly: A deep-reinforcement-learning approach. {\em Applied Sciences}. \textbf{10}, 6923 (2020)
\bibitem{wang2020learning}Wang, C., Zhang, Q., Tian, Q., Li, S., Wang, X., Lane, D., Petillot, Y. \& Wang, S. Learning mobile manipulation through deep reinforcement learning. {\em Sensors}. \textbf{20}, 939 (2020)
\bibitem{kindle2020whole}Kindle, J., Furrer, F., Novkovic, T., Chung, J., Siegwart, R. \& Nieto, J. Whole-body control of a mobile manipulator using end-to-end reinforcement learning. {\em ArXiv Preprint ArXiv:2003.02637}. (2020)
\bibitem{honerkamp2021learning}Honerkamp, D., Welschehold, T. \& Valada, A. Learning kinematic feasibility for mobile manipulation through deep reinforcement learning. {\em IEEE Robotics And Automation Letters}. \textbf{6}, 6289-6296 (2021)
\bibitem{sun2022fully}Sun, C., Orbik, J., Devin, C., Yang, B., Gupta, A., Berseth, G. \& Levine, S. Fully autonomous real-world reinforcement learning with applications to mobile manipulation. {\em Conference On Robot Learning}. pp. 308-319 (2022)
\bibitem{liu2021deep}Liu, R., Nageotte, F., Zanne, P., Mathelin, M. \& Dresp-Langley, B. Deep reinforcement learning for the control of robotic manipulation: a focussed mini-review. {\em Robotics}. \textbf{10}, 22 (2021)
\bibitem{ibarz2021train}Ibarz, J., Tan, J., Finn, C., Kalakrishnan, M., Pastor, P. \& Levine, S. How to train your robot with deep reinforcement learning: lessons we have learned. {\em The International Journal Of Robotics Research}. \textbf{40}, 698-721 (2021)
\bibitem{gu2020deep}Gu, S., Holly, E., Lillicrap, T. \& Levine, S. Deep reinforcement learning for robotic manipulation with asynchronous off-policy updates. {\em 2017 IEEE International Conference On Robotics And Automation (ICRA)}. pp. 3389-3396 (2017)
\bibitem{nemec2017door}Nemec, B., Žlajpah, L. \& Ude, A. Door opening by joining reinforcement learning and intelligent control. {\em 2017 18th International Conference On Advanced Robotics (ICAR)}. pp. 222-228 (2017)
\bibitem{Door_gym}Urakami, Y., Hodgkinson, A., Carlin, C., Leu, R., Rigazio, L. \& Abbeel, P. Doorgym: A scalable door opening environment and baseline agent. {\em ArXiv Preprint ArXiv:1908.01887}. (2019)
\bibitem{wang2022research}Wang, Y., Wang, L. \& Zhao, Y. Research on Door Opening Operation of Mobile Robotic Arm Based on Reinforcement Learning. {\em Applied Sciences}. \textbf{12}, 5204 (2022)
\bibitem{scirobotics}Ito, H., Yamamoto, K., Mori, H. \& Ogata, T. Efficient multitask learning with an embodied predictive model for door opening and entry with whole-body control. {\em Science Robotics}. \textbf{7}, eaax8177 (2022)
\bibitem{liu2021yolactedge}Liu, H., Soto, R., Xiao, F. \& Lee, Y. Yolactedge: Real-time instance segmentation on the edge. {\em 2021 IEEE International Conference On Robotics And Automation (ICRA)}. pp. 9579-9585 (2021)
\bibitem{lai2022stratified}Lai, X., Liu, J., Jiang, L., Wang, L., Zhao, H., Liu, S., Qi, X. \& Jia, J. Stratified transformer for 3d point cloud segmentation. {\em Proceedings Of The IEEE/CVF Conference On Computer Vision And Pattern Recognition}. pp. 8500-8509 (2022)
\bibitem{fischler1981random}Fischler, M. \& Bolles, R. Random sample consensus: a paradigm for model fitting with applications to image analysis and automated cartography. {\em Communications Of The ACM}. \textbf{24}, 381-395 (1981)
\bibitem{SAC}Haarnoja, T., Zhou, A., Abbeel, P. \& Levine, S. Soft actor-critic: Off-policy maximum entropy deep reinforcement learning with a stochastic actor. {\em International Conference On Machine Learning}. pp. 1861-1870 (2018)
\bibitem{fuchs2020super}Fuchs, F., Song, Y., Kaufmann, E., Scaramuzza, D. \& Duerr, P. Super-Human Performance in Gran Turismo Sport Using Deep Reinforcement Learning. {\em ArXiv Preprint ArXiv:2008.07971}. (2020)
\bibitem{seong2021learning}Seong, H., Jung, C., Lee, S. \& Shim, D. Learning to drive at unsignalized intersections using attention-based deep reinforcement learning. {\em 2021 IEEE International Intelligent Transportation Systems Conference (ITSC)}. pp. 559-566 (2021)
\bibitem{yang2019control}Yang, C., Yang, J., Wang, X. \& Liang, B. Control of space flexible manipulator using soft actor-critic and random network distillation. {\em 2019 IEEE International Conference On Robotics And Biomimetics (ROBIO)}. pp. 3019-3024 (2019)
\bibitem{prianto2020path}Prianto, E., Kim, M., Park, J., Bae, J. \& Kim, J. Path planning for multi-arm manipulators using deep reinforcement learning: Soft actor–critic with hindsight experience replay. {\em Sensors}. \textbf{20}, 5911 (2020)
\bibitem{haarnoja2018softapplication}Haarnoja, T., Zhou, A., Hartikainen, K., Tucker, G., Ha, S., Tan, J., Kumar, V., Zhu, H., Gupta, A., Abbeel, P. \& Others Soft actor-critic algorithms and applications. {\em ArXiv Preprint ArXiv:1812.05905}. (2018)
\bibitem{fujimoto2018addressing}Fujimoto, S., Hoof, H. \& Meger, D. Addressing function approximation error in actor-critic methods. {\em International Conference On Machine Learning}. pp. 1587-1596 (2018)


\end{thebibliography}


\end{document}